\documentclass[10pt,twocolumn,letterpaper]{article}

\usepackage{cvpr}              % To produce the CAMERA-READY version
\usepackage{algorithm}
\usepackage{multirow} % table related
\usepackage{bbding} % table related
\usepackage[table,xcdraw]{xcolor}
\usepackage[accsupp]{axessibility}  % Improves PDF readability for those with disabilities.

\usepackage{array}
\usepackage{makecell}
\usepackage{pifont}
\newcommand{\cmark}{\ding{51}} % ✓
\newcommand{\xmark}{\ding{55}} % ✗

\usepackage{xspace}
\newcommand{\method}{VerseCrafter\xspace}

\definecolor{cvprblue}{rgb}{0.21,0.49,0.74}
\usepackage[pagebackref,breaklinks,colorlinks,allcolors=cvprblue]{hyperref}

\title{\method: Dynamic Realistic Video World Model with 4D Geometric Control}

\author{
   \vspace*{0.1cm}
   \!Sixiao Zheng$^{1,2}$\ , \  Minghao Yin$^{3}$,\  Wenbo Hu$^4$$^\dag$,\  Xiaoyu Li$^{4}$,\  Ying Shan$^4$,\  Yanwei Fu$^{1,2}$$^\dag$\\
   \vspace*{0.1cm}
   {$^1$}Fudan University {$^2$}Shanghai Innovation Institute {$^3$}HKU {$^4$}ARC Lab, Tencent PCG \\
   \small Project Page: \url{https://sixiaozheng.github.io/VerseCrafter_page/}
}

\begin{document}

\twocolumn[{
      \renewcommand\twocolumn[1][]{#1}%
      \maketitle
      \begin{figure}[H]
        \hsize=\textwidth
        \centering
        \includegraphics[width=6.3 in]{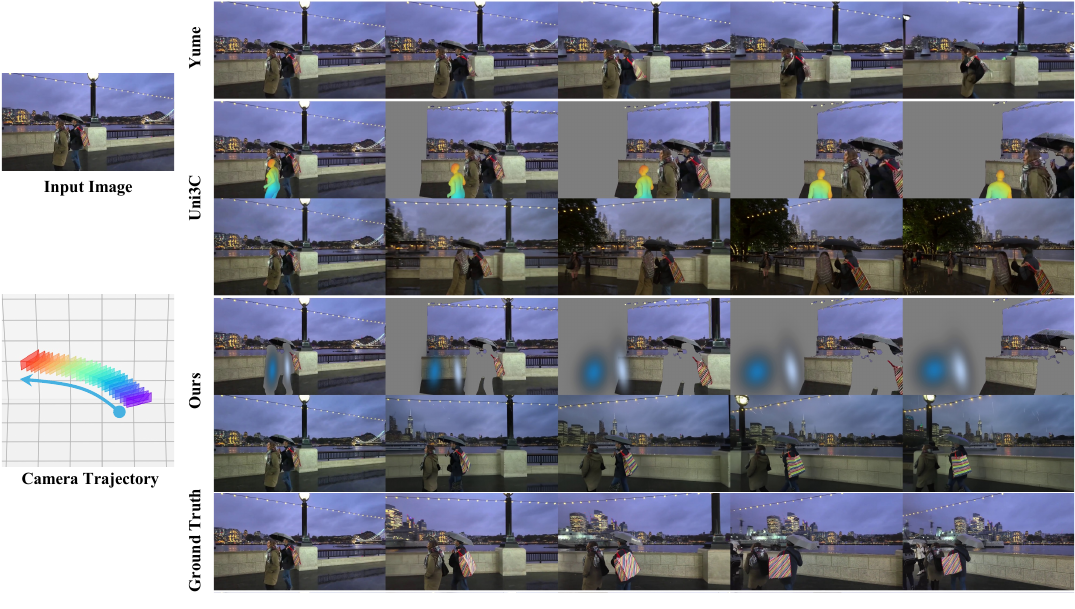}
        \caption{
        \textbf{\method} enables precise control of camera motion and multi-object motion via a 4D Geometric Control representation built from a static background point cloud and per-object 3D Gaussian trajectories, producing videos that better follow the desired motion than Yume~\cite{mao2025yume} and Uni3C~\cite{cao2025uni3c} and more closely match the ground-truth video.
        \label{Figure:Teaser} }
      \end{figure}
    }]

\renewcommand{\thefootnote}{}
\footnotetext[0]{ $^\dag$Corresponding authors.}

% \maketitle
\begin{abstract}
Video world models aim to simulate dynamic, real-world environments, yet existing methods struggle to provide unified and precise control over camera and multi-object motion, as videos inherently capture dynamics in the projected 2D image plane.
To bridge this gap, we introduce \textbf{\method}, a geometry-driven video world model that generates dynamic, realistic videos from a unified 4D geometric world state.
Our approach is centered on a novel \textbf{4D Geometric Control} representation, which encodes the world state as a static background point cloud and per-object 3D Gaussian trajectories.
This representation captures each object's motion path and probabilistic 3D occupancy over time, providing a flexible, category-agnostic alternative to rigid bounding boxes and parametric models.
We render 4D Geometric Control into 4D control maps for a pretrained video diffusion model, enabling high-fidelity, view-consistent video generation that faithfully follows the specified dynamics.
To enable training at scale, we develop an automatic data engine and construct \textbf{VerseControl4D}, a real-world dataset of 35K training samples with automatically derived prompts and rendered 4D control maps.
Extensive experiments show that \method achieves superior visual quality and more accurate control over camera and multi-object motion than prior methods.
\end{abstract}
    
\section{Introduction}
\label{sec:intro}

Video world models learn to simulate world dynamics by generating future frame sequences conditioned on past observations and control signals, such as actions or camera trajectories~\cite{lecun2022path,ha2018world,chen2025deepverse,huang2025voyager,mao2025yume}.
They provide a unified interface for visual prediction~\cite{hafner2023mastering}, navigation~\cite{bar2025navigation}, and manipulation~\cite{ferraro2025focus}.
However, the reliance on video introduces a fundamental challenge: while an ideal world model should simulate the full 4D spatiotemporal space to reflect our physical reality, videos inherently capture dynamics in the projected 2D image plane.

To bridge this gap, recent works introduce camera control into video generation through explicit 3D geometry~\cite{yu2024viewcrafter,cao2025uni3c,zheng2025vidcraft3}, implicit pose embeddings~\cite{li2025vmem}, or learned movement embeddings~\cite{bruce2024genie,parkerholder2024genie2,che2024gamegen}.
However, these methods are often limited to static scenes or leave multi-object motion uncontrolled.
Existing approaches typically rely on 2D cues such as point trajectories~\cite{wang2024motionctrl}, optical flow~\cite{liao2025motionagent}, masks~\cite{zhang2025motionpro}, or bounding boxes~\cite{wang2024boximator}, which lack 3D awareness and often fail under large viewpoint changes.
More advanced 3D-aware methods use depth maps~\cite{zhang2025i2v3d}, sparse 3D trajectories~\cite{chen2025perception}, 3D bounding boxes~\cite{wang2025cinemaster}, or parametric human models like SMPL-X~\cite{cao2025uni3c} to align camera and object motion in 3D space.
Nevertheless, these control representations remain inadequate for modeling multi-object dynamics as a unified, compact, and editable 4D geometric scene state in a shared world coordinate frame.
For instance, sparse trajectories are often noisy and incomplete, 3D bounding boxes impose rigid constraints ill-suited to natural objects, and SMPL-X representations are category-limited.
Furthermore, several existing works focus on synthetic game environments~\cite{huang2025voyager,yang2025matrix,yu2025gamefactory}, where precise annotations are available for training. 
However, controllable modeling of complex, realistic 4D scenes with multi-object motion remains underexplored.

Thus we propose \emph{\method}, a realistic, dynamic video world model that enables precise control of camera and multi-object motion within a unified 4D geometric world state, as shown in Fig.~\ref{Figure:Teaser}.
At the core of VerseCrafter is our \emph{4D Geometric Control} representation, which represents the scene state as a static background point cloud for scene geometry and per-object 3D Gaussian trajectories to capture object dynamics.
Each 3D Gaussian trajectory models an object's probabilistic 3D occupancy over time: its mean defines the motion path, while its covariance captures the object's spatial extent and orientation.
This probabilistic formulation provides a soft, flexible, and category-agnostic way to model diverse object shapes and motions, overcoming the limitations of rigid 3D bounding boxes or category-specific parametric models.
Crucially, the background point cloud and per-object 3D Gaussian trajectories share a common world coordinate frame, enabling coherent and unified control over both camera and object motion.

By rendering 4D Geometric Control into multi-channel 4D control maps, we condition a frozen Wan2.1-14B video diffusion backbone~\cite{wan2025wan} via a lightweight GeoAdapter, an adapter-style branch inspired by ControlNet~\cite{zhang2023adding}.
This conditioning enables the generation of high-fidelity videos that faithfully reflect the explicit 4D geometric world state.
Unlike 2D control signals, our 4D Geometric Control is inherently 3D-aware, making it naturally more view-consistent and robust to occlusions, and thus a more effective and reliable interface for video world modeling.
Training VerseCrafter requires large-scale paired data of real-world videos and corresponding 4D geometric control.
To this end, we construct \textit{VerseControl4D}, a real-world video dataset with automatically derived prompts and rendered 4D control maps.
This dataset supports large-scale training on diverse real-world videos.

Our contributions are threefold:
\begin{itemize}
    \item We introduce a novel \textit{\textbf{4D Geometric Control}} representation that unifies camera and multi-object motion in a shared world coordinate frame. By using 3D Gaussian trajectories, it provides a flexible and category-agnostic way to control object dynamics, overcoming the limitations of rigid, category-specific models.
    
    \item We present \textit{\textbf{{\method}}}, a geometry-driven video world model that leverages 4D Geometric Control for precise control over camera and multi-object motion. 
    This enables the generation of high-fidelity, view-consistent videos that accurately follow complex 4D controls.

    \item We construct \textit{\textbf{VerseControl4D}}, a real-world dataset with automatically derived prompts and rendered 4D control maps, with 35K training samples. This addresses a key data bottleneck and supports large-scale training on diverse real-world videos.
\end{itemize}

\vspace{-5pt}
\section{Related Works}
\label{sec:related_works}
\vspace{-5pt}

\noindent \textbf{Video World Models}.
World models learn environment dynamics from observations by predicting future states for  simulation, planning, and control \cite{ha2018world,hafner2019learning,lecun2022path}.
Early visual world models adopt recurrent and latent-variable architectures \cite{finn2016unsupervised,villegas2019high,chiappa2017recurrent,ha2018recurrent,menapace2021playable,oh2015action}, while recent approaches use transformer and diffusion backbones to roll out realistic videos conditioned on actions, text, or camera trajectories \cite{bruce2024genie,parkerholder2024genie2,genie3,wan2025wan,yang2024cogvideox,kong2024hunyuanvideo,agarwal2025cosmos,alhaija2025cosmos,he2025matrix,hu2023gaia,che2024gamegen,yu2025gamefactory,decart2024oasis,xiang2024pandora,li2025hunyuan}, and further extend temporal horizons with memories or long-sequence models \cite{po2025long,wu2025video,li2025vmem}.
Geometry-aware works such as DeepVerse \cite{chen2025deepverse}, Voyager \cite{huang2025voyager}, and Yume \cite{mao2025yume} incorporate 3D geometry to support 4D video generation and world exploration, but are controlled via text, actions, or camera tokens and do not expose a compact, editable 4D geometric state for real-world multi-object dynamics. 
In contrast, VerseCrafter learns a geometry-driven mapping from 4D Geometric Control to dynamic,  realistic videos, enabling disentangled control over camera and multi-object motion.

\noindent \textbf{3D World Generation}.
Recent work leverages powerful 2D generative priors to synthesize explorable 3D environments from text, images, or videos \cite{zhang2024towards,li2024dreamscene}. 
Early methods mainly focus on object-level or single-scene generation \cite{hong2023lrm,zhang2024gs,zhang2024clay,poole2022dreamfusion,cohen2023set,zhang2025scene}, distilling image diffusion models \cite{rombach2022high} into NeRFs \cite{mildenhall2021nerf}, implicit fields, meshes, or 3D Gaussian splats \cite{kerbl20233d}, or optimizing scene geometry from multi-view or panoramic observations \cite{sargent2024zeronvs,yu2023long,chung2023luciddreamer,yu2024wonderjourney,yu2025wonderworld}. 
More recent approaches scale up to navigable 3D worlds \cite{bar2025navigation}, combining depth estimation \cite{yang2024depth}, camera-guided video diffusion, iterative inpainting, and panoramic inputs to construct room- or city-scale Gaussian scenes for exploration \cite{schneider2025worldexplorer,team2025hunyuanworld,yang2025matrix,chen2025flexworld,li2025flashworld,liu2025worldmirror,wang2025evoworld,zhu2025aether,lu2024genex}. 
However, these pipelines largely model static, synthetic-like scenes and provide limited explicit control over real-world multi-object dynamics. In contrast, VerseCrafter operates on real-world videos and represents the scene with a static background point cloud and per-object 3D Gaussian trajectories, forming an explicit 4D geometric scene state for geometry-consistent dynamic video generation.

\noindent \textbf{Controllable Video Generation}.
Controllable video generation aims to steer camera and object motion via conditioning signals. 
Camera-controlled models \cite{zheng2024cami2v,bahmani2025ac3d,sun2024dimensionx,bahmani2024vd3d,kuang2024collaborative,he2025cameractrl,xu2024camco,li2025realcam} such as MotionCtrl \cite{wang2024motionctrl} and CameraCtrl \cite{he2024cameractrl} inject camera extrinsics, Pl\"ucker-style encodings, or 3D priors \cite{xiao2024trajectory,hou2024training,wang2025epic,zhang2025i2v3d,ren2025gen3c,popov2025camctrl3d,yu2024viewcrafter,gu2025diffusion,cao2025uni3c,feng2024i2vcontrolcamera,zheng2025vidcraft3} into video diffusion models for viewpoint control, but mostly assume static or weakly dynamic scenes. 
Object motion \cite{tanveer2024motionbridge,wang2023videocomposer,xu2024motion,li2024image,mou2024revideo,niu2025mofa,shi2024motion,qiu2024freetraj,ma2024trailblazer,li2025magicmotion,zhou2024trackgo,zhang2024tora,wang2024levitor,wan2024dragentity,he2024mojito,pandey2024motion,wang2024objctrl,fu20243dtrajmaster,liang2025realismotion,shuai2025free,burgert2025go,yang2024direct,yin2023dragnuwa,wu2024motionbooth} is typically controlled using 2D cues (bounding boxes, masks, trajectories, strokes, optical flow) as in Boximator \cite{wang2024boximator}, DragAnything \cite{wu2025draganything}, and MotionCanvas \cite{xing2025motioncanvas}, or with more 3D-aware signals such as depth maps, sparse 3D trajectories, 3D boxes, or SMPL-X bodies in I2V3D \cite{zhang2025i2v3d}, Uni3C \cite{cao2025uni3c}, CineMaster \cite{wang2025cinemaster}, Perception-as-Control \cite{chen2025perception}, and LongVie \cite{gao2025longvie}. 
While these methods improve controllability, 2D controls remain view-dependent and fragile under large camera changes, and many 3D controls are category-specific, rigid, or tied to reconstruction-heavy pipelines. 
Recent approaches \cite{yang2024direct,geng2024motion,liao2025motionagent,xing2025motioncanvas,wang2025cinemaster,feng2024i2vcontrol,wang2024motionctrl,zheng2025vidcraft3,chen2025perception} begin to jointly control camera and object motion, but their control spaces are still fragmented rather than a unified, compact world state. 
VerseCrafter instead introduces \emph{4D Geometric Control}: a compact, category-agnostic 4D geometric scene state where a static background point cloud and per-object 3D Gaussian trajectories in a shared world coordinate frame jointly drive camera and multi-object motion.

\begin{figure*}
    \centering
    \includegraphics[width=0.93\linewidth]{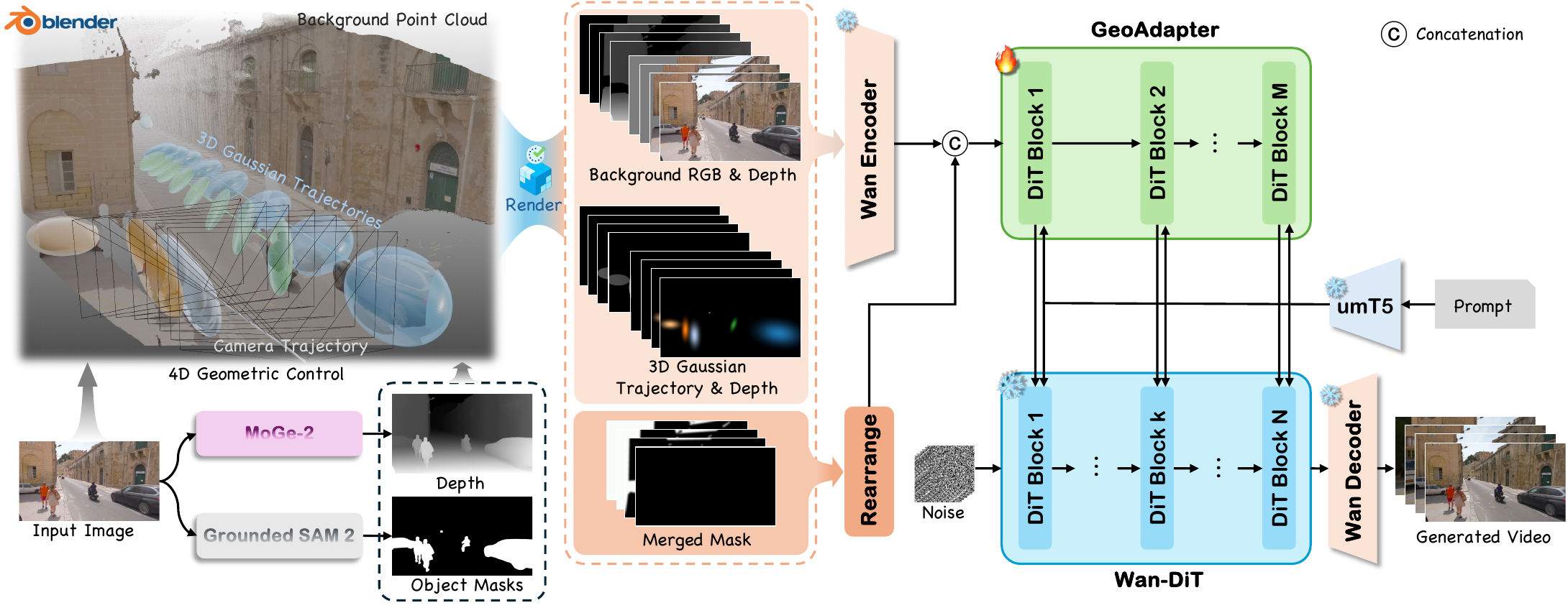}
    \caption{\textbf{Framework of VerseCrafter.}
    Given an input image and a text prompt, we estimate depth and obtain user-specified object masks to construct \textit{4D Geometric Control} consisting of a static background point cloud and per-object 3D Gaussian trajectories in a shared world coordinate frame.
    A camera trajectory is specified in the shared frame, and together with the 4D Geometric Control, rendered into per-frame background RGB/depth, 3D Gaussian trajectory RGB/depth, and a soft merged mask, forming multi-channel 4D control maps.
    The 4D control maps are encoded and fed into the proposed GeoAdapter, which conditions a frozen Wan2.1-14B backbone together with text embeddings from umT5, enabling geometry-consistent video generation with precise control over camera and multi-object motion.   \label{fig:framework}}
\end{figure*}

\section{Method}
\label{sec:method}

We propose \textbf{VerseCrafter}, a geometry-driven video world model that generates dynamic, realistic videos from an explicit 4D geometric scene state while enabling disentangled control over camera and multi-object motion. Our framework has two key components: (i) a unified \emph{4D Geometric Control} representation (Sec.~\ref{sec:4d_control}), which represents the 4D geometric scene state in a shared world coordinate frame, and (ii) a lightweight \emph{GeoAdapter} (Sec.~\ref{sec:architecture}), which injects encoded 4D control maps into a frozen Wan2.1-14B backbone while preserving its strong visual prior. Given an input image and a prompt, we construct 4D Geometric Control as a static background point cloud and per-object 3D Gaussian trajectories, specify a camera trajectory in the shared frame, render them into 4D control maps, and feed these maps into GeoAdapter to generate dynamic, realistic videos.

\subsection{4D Geometric Control}
\label{sec:4d_control}

We represent each scene as an explicit 4D geometric scene state, which we term \emph{4D Geometric Control}. This editable representation consists of a static background point cloud $P^{\text{bg}}$ and per-object 3D Gaussian trajectories $\{\mathcal{G}_o^t\}$, all defined in a shared world coordinate frame.

\noindent\textbf{Background point cloud.}
As shown in Fig.~\ref{fig:framework}, we start from the input image, estimate monocular depth and camera intrinsics $\mathbf{K}$ using MoGe-2~\cite{wang2025moge}, and obtain object masks $\{M_o\}$ with Grounded SAM2~\cite{ren2024grounded}, where the user selects one or more objects to control via text prompts or clicks. We take the input view as the reference world coordinate frame, so that the reference camera pose is given by $\mathbf{R}_1=\mathbf{I}$ and $\mathbf{t}_1=\mathbf{0}$. Each pixel $\mathbf{u}=(u,v,1)^\top$ with depth $D_1(\mathbf{u})$ is then back-projected as
\begin{equation}
\mathbf{p}(\mathbf{u}) = \mathbf{R}_1^\top \big(D_1(\mathbf{u}) \mathbf{K}^{-1} \mathbf{u} - \mathbf{t}_1 \big).
\end{equation}
We use the object masks to partition the reconstructed point cloud into per-object point clouds
\begin{equation}
P_o = \big\{\mathbf{x}_{o,k} \,\big|\, \mathbf{x}_{o,k} = \mathbf{p}(\mathbf{u}_k),\ \mathbf{u}_k \in M_o \big\},
\end{equation}
and a static background point cloud
\begin{equation}
P^{\text{bg}} = \big\{\mathbf{p}(\mathbf{u}) \,\big|\, \mathbf{u} \notin \bigcup_o M_o \big\} = \{\mathbf{p}_i\}_{i=1}^{N_{\text{bg}}}.
\end{equation}
During generation, the background at frame $t$ is rendered from $P^{\text{bg}}$ under the camera pose, so viewpoint changes are realized as rigid camera motion in a fixed 3D world rather than by hallucinating a new background at every frame.

\noindent\textbf{3D Gaussian trajectories.}
A single 3D Gaussian $\mathcal{G}_o(\mathbf{x}) = \mathcal{N}(\mathbf{x}\mid \boldsymbol{\mu}_o,\mathbf{\Sigma}_o)$ in the world coordinate frame compactly encodes an object’s position (through $\boldsymbol{\mu}_o$), approximate shape and size (through the eigenvalues of $\mathbf{\Sigma}_o$), and orientation (through the eigenvectors of $\mathbf{\Sigma}_o$).
\emph{A 3D Gaussian trajectory for an object $o$} is then defined as a sequence of Gaussians
\begin{equation}
\{\mathcal{G}_o^t\}_{t=1}^T,\quad 
\mathcal{G}_o^t(\mathbf{x}) = \mathcal{N}(\mathbf{x}\mid \boldsymbol{\mu}_o^t,\mathbf{\Sigma}_o^t),
\end{equation}
whose means $\{\boldsymbol{\mu}_o^t\}$ trace the motion path in 3D, while the covariances $\{\mathbf{\Sigma}_o^t\}$ capture how the object’s spatial extent and orientation evolve over time. This probabilistic formulation describes the object’s 3D occupancy in a soft, continuous manner and yields a compact control space that is more flexible than rigid 3D bounding boxes and more category-agnostic than parametric body models.

To initialize the trajectory for each controllable object $o$, we fit a full-covariance Gaussian to its point cloud $P_o$ obtained in the previous step:
\begin{equation}
\boldsymbol{\mu}_o = \frac{1}{N_o}\sum_k \mathbf{x}_{o,k}, 
\mathbf{\Sigma}_o = \frac{1}{N_o}\sum_k (\mathbf{x}_{o,k} - \boldsymbol{\mu}_o)(\mathbf{x}_{o,k} - \boldsymbol{\mu}_o)^\top,
\end{equation}
which gives an initial Gaussian $\mathcal{G}_o(\mathbf{x})$

The low-dimensional parameters $\{\boldsymbol{\mu}_o^t,\mathbf{\Sigma}_o^t\}$ naturally support flexible, user-driven editing. In practice, we convert each $\mathcal{G}_o^t$ into an ellipsoid mesh for visualization in a 3D editor such as Blender, and let the user specify or refine the trajectory by dragging and keyframing this ellipsoid in world coordinate space. The edited poses and shapes are mapped back to $\{\boldsymbol{\mu}_o^t,\mathbf{\Sigma}_o^t\}$ as control signals. The ellipsoids are only a user interface; all conditioning maps used by model are rendered directly from the underlying 3D Gaussians.

\noindent\textbf{Rendering 4D control maps.}
Given our 4D Geometric Control, we render per-frame 4D control maps in the target camera views. For each frame $t$, we generate three types of maps:
(i) background RGB/depth, $\text{RGB}^{\text{bg}}_t$ and $\text{Depth}^{\text{bg}}_t$, by projecting the static background point cloud $P^{\text{bg}}$ under the camera pose $(\mathbf{R}_t,\mathbf{t}_t)$;
(ii) 3D Gaussian trajectory RGB/depth, $\text{RGB}^{\text{traj}}_t$ and $\text{Depth}^{\text{traj}}_t$, by projecting the per-object Gaussians $\{\mathcal{G}_o^t\}$ into soft elliptical footprints and taking depth from the corresponding ellipsoid surfaces;
and (iii) a soft merged mask $M_t$, used as a control mask, that indicates regions where the diffusion model should synthesize or overwrite content, obtained by inverting background visibility and merging it with the projected 3D Gaussian footprints, followed by Gaussian smoothing. For the first frame $t=1$, we replace $\text{RGB}^{\text{bg}}_1$ with  input image and set $M_1 = 0$, so that the first frame is preserved and only future frames are modified.
Background and 3D Gaussian maps share the same 4D geometric scene state but are rendered through \emph{decoupled} channels, disentangling camera motion from object motion while preserving geometric consistency.

\subsection{VerseCrafter Architecture}
\label{sec:architecture}

\noindent\textbf{Backbone.}
We adopt Wan2.1-14B~\cite{wan2025wan} as a frozen latent video diffusion backbone with a Wan Encoder, a Wan-DiT denoiser and a Wan Decoder.
VerseCrafter treats Wan2.1 as a generic video prior: we do not change its architecture or weights, and instead attach a lightweight geometric adapter that conditions the backbone on rendered 4D control maps.

\noindent\textbf{GeoAdapter.}
We take the rendered background maps and 3D Gaussian trajectory maps,
$\text{RGB}^{\text{bg}}$, $\text{Depth}^{\text{bg}}$, $\text{RGB}^{\text{traj}}$, $\text{Depth}^{\text{traj}}$, together with the soft merged mask $M$.
The four RGB/depth maps are encoded by the same Wan Encoder, while $M$ is reshaped and interpolated to the latent resolution, following the practice in~\cite{jiang2025vace,wan2025wan}.
The encoded RGB/depth maps and the processed mask are concatenated channel-wise to form a spatio-temporal geometry tensor.
GeoAdapter is a lightweight DiT-style branch that operates on this geometry tensor.
It shares the same hidden dimensionality as the Wan-DiT blocks, but uses far fewer layers.
We interleave GeoAdapter blocks with the frozen Wan-DiT: every $k$-th DiT block in Wan-DiT is paired with a GeoAdapter block whose output is linearly projected and added to the corresponding DiT block as a residual modulation.
Text prompts are encoded by umT5~\cite{chung2023unimax} into text embeddings, which are injected into both Wan-DiT and GeoAdapter blocks through the same text-conditioning interfaces.
This adapter-based conditioning injects 4D geometric information into Wan 2.1 with only a small number of extra parameters, while keeping all backbone weights fixed.

\noindent\textbf{Inference.}
At inference time, VerseCrafter supports camera-only, object-only, and joint control within a unified framework.
For \emph{camera-only} control, we render the background control maps, while setting the 3D Gaussian trajectory RGB/depth maps to zero. The merged mask may still be nonzero due to viewpoint changes.
For \emph{object-only} control, we keep the camera pose fixed, render static background control maps from $P^{\text{bg}}$, and render the 3D Gaussian trajectory maps to control object motion.
For \emph{joint} control, both background and trajectory control maps are rendered from the same 4D geometric scene state, enabling coordinated and geometry-consistent control over camera and multi-object motion.

\section{VerseControl4D Dataset}
\label{sec:data}
\begin{figure}[t]
    \centering
    \includegraphics[width=1\linewidth]{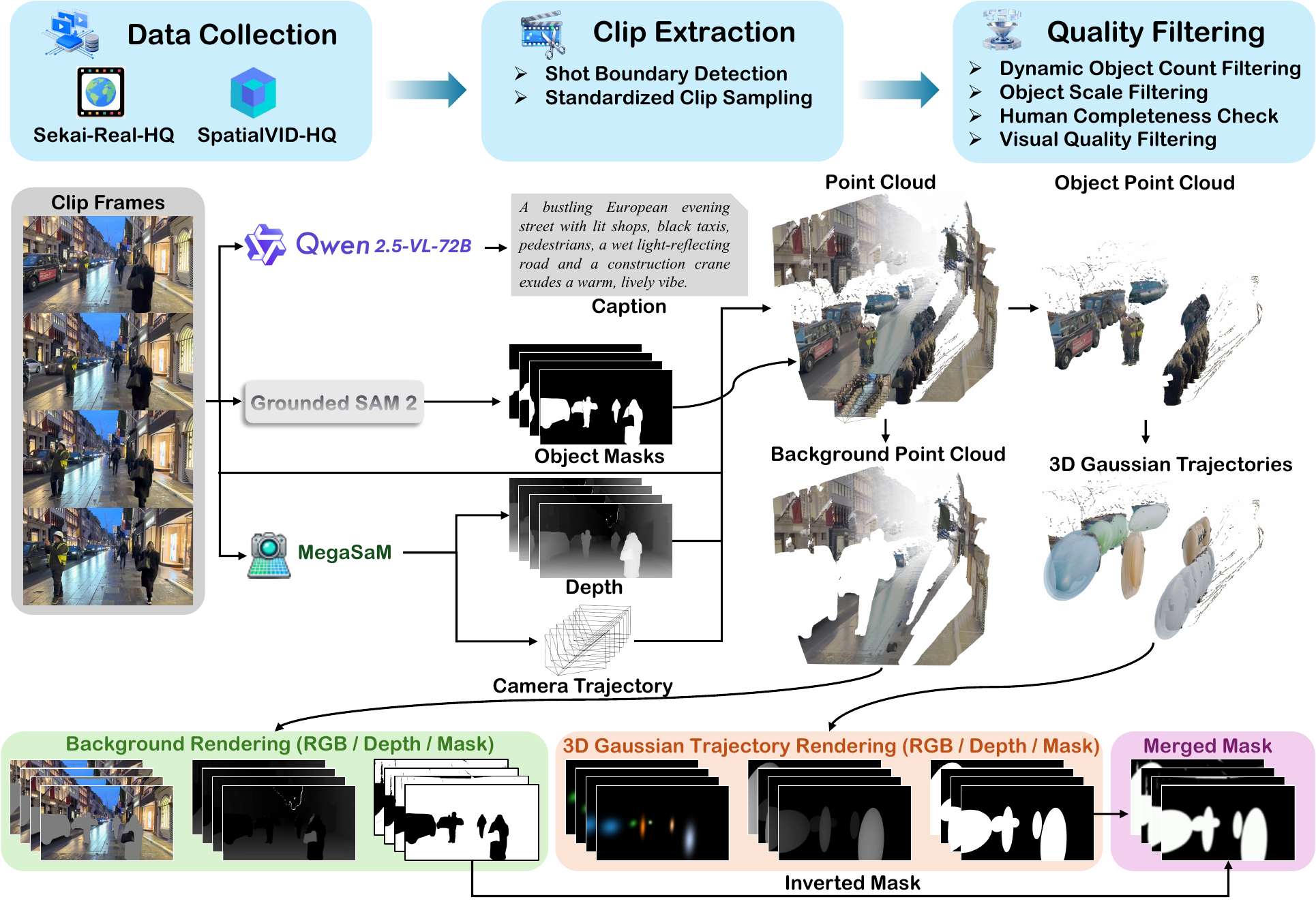}
    \caption{\textbf{Construction pipeline of VerseControl4D.} Starting from Sekai-Real-HQ and SpatialVID-HQ, we extract 81-frame clips and apply quality filtering. For each retained clip, Qwen2.5-VL-72B, Grounded-SAM2, and MegaSAM provide captions, object masks, depth, and camera trajectory, which are lifted into background/object point clouds, from which 3D Gaussian trajectories are fitted, and then rendered into background/trajectory maps and a soft merged mask that constitute our 4D control maps.
    \label{fig:data_pipeline} }
\end{figure}

To train and evaluate VerseCrafter on complex real-world scenes with 4D Geometric Control, we construct \textbf{VerseControl4D}, a real-world dataset with automatically derived prompts and rendered 4D control maps. As shown in Fig.~\ref{fig:data_pipeline}, VerseControl4D is built through four stages: data collection, clip extraction, quality filtering, and data annotation.

\noindent\textbf{Data collection.}
VerseControl4D is built from two recent world-exploration datasets, Sekai-Real-HQ~\cite{li2025sekai} and SpatialVID-HQ~\cite{wang2025spatialvid}, which provide long in-the-wild videos with diverse outdoor and urban scenes, camera poses, and captions, but lack object-motion labels.
We use their high-resolution videos as the raw video pool for constructing our 4D geometric control annotations.

\noindent\textbf{Clip extraction.}
We apply PySceneDetect to detect shots in the videos.
For each shot longer than 81 frames, we uniformly sample an 81-frame sub-clip and discard shorter shots, matching the default temporal length used by the Wan2.1 backbone.

\noindent\textbf{Quality filtering.}
We apply an object-centric filtering pipeline to retain clips with clean geometry and controllable foreground.
Using Grounded-SAM2 with prompts such as \emph{``person . human . car . animal''}, we first obtain object masks on the first frame, and keep only clips whose controllable object count lies in $[1, 6]$.
We then discard clips where any object mask covers more than $20\%$ of the image area.
For human instances, we further remove clips whose masks touch image borders or whose aspect ratios fall outside $[2, 4]$, as these typically correspond to severely truncated pedestrians.
Finally, we apply visual-quality filtering based on aesthetic and luminance scores to exclude blurry or over-/under-exposed clips, yielding a set of visually clean, geometrically reliable videos.

\begin{figure*}
    \centering
    \includegraphics[width=1\linewidth]{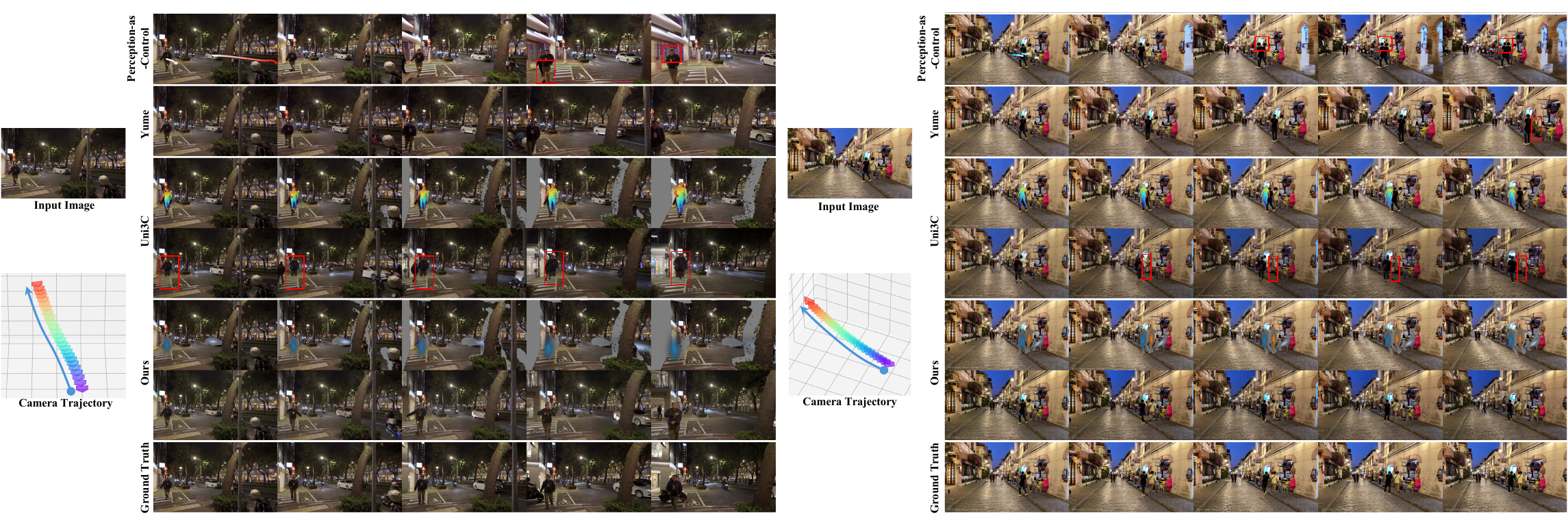}
    \caption{\textbf{Qualitative comparison of joint camera and object motion control.}
    Perception-as-Control and Uni3C exhibit noticeable human deformation, while Yume roughly follows the text-described motion but lacks precise camera control. Uni3C is also limited to single human. In contrast, VerseCrafter more faithfully follows both the camera trajectory and multi-object motion while maintaining sharp appearance and geometrically consistent backgrounds.
\label{fig:joint_control_vis} }
\vspace{-5pt}
\end{figure*}

\begin{figure*}
    \centering
    \includegraphics[width=1\linewidth]{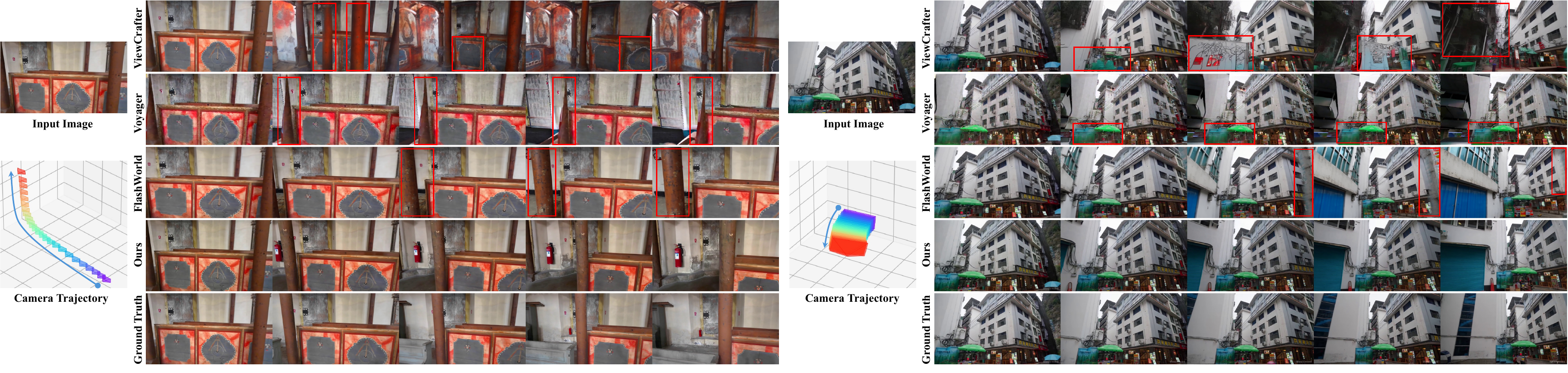}
    \caption{\textbf{Qualitative comparison of camera-only motion control on static scenes.} 
    ViewCrafter and Voyager exhibit distorted facades, drifting structures, or inaccurate camera motion, while FlashWorld tends to produce blurred scene boundaries and imprecise camera motion. In contrast, VerseCrafter better follows the target camera trajectory while preserving sharp details and globally consistent 3D geometry.
\label{fig:camera_only_vis}}
\vspace{-5pt}
\end{figure*}

\setlength{\tabcolsep}{1pt}
\begin{table*}[]
\caption{\textbf{Joint camera and object motion control on VerseControl4D.}
We report VBench-I2V scores and 3D control metrics (RotErr, TransErr, ObjMC).
VerseCrafter achieves the best overall video quality and the most accurate joint control of camera and object motion.}
\label{tab:joint_control}
\scriptsize
\centering
\begin{tabular}{l|ccccccccc|ccc}
\toprule
                               & \textbf{\begin{tabular}[c]{@{}c@{}}Overall \\ Score\end{tabular}$\uparrow$} & \textbf{\begin{tabular}[c]{@{}c@{}}Imaging \\ Quality\end{tabular}$\uparrow$} & \textbf{\begin{tabular}[c]{@{}c@{}}Aesthetic \\ Quality\end{tabular}$\uparrow$} & \textbf{\begin{tabular}[c]{@{}c@{}}Dynamic \\ Degree\end{tabular}$\uparrow$} & \textbf{\begin{tabular}[c]{@{}c@{}}Motion \\ Smoothness\end{tabular}$\uparrow$} & \textbf{\begin{tabular}[c]{@{}c@{}}Background \\ Consistency\end{tabular}$\uparrow$} & \textbf{\begin{tabular}[c]{@{}c@{}}Subject\\  Consistency\end{tabular}$\uparrow$} & \textbf{\begin{tabular}[c]{@{}c@{}}I2V \\ Background\end{tabular}$\uparrow$} & \textbf{\begin{tabular}[c]{@{}c@{}}I2V \\ Subject\end{tabular}$\uparrow$} & \textbf{RotErr$\downarrow$} & \textbf{TransErr$\downarrow$} & \textbf{ObjMC$\downarrow$} \\
                                   \midrule
\textbf{Perception-as-Control \cite{chen2025perception}} & 83.66                                                             & 66.81                                                               & 53.34                                                                 & 73.91                                                              & 96.89                                                                 & 93.19                                                                      & 94.02                                                                   & 96.35                                                              & 94.78                                                           & 5.006           & 8.767             & 6.556          \\
\textbf{Yume \cite{mao2025yume}}                  & 85.47                                                             & 71.16                                                               & 52.39                                                                 & 72.24                                                              & \textbf{98.96}                                                        & 95.66                                                                      & 96.43                                                                   & 98.51                                                              & 98.39                                                           & 7.560           & 8.735             & 7.959          \\
\textbf{Uni3C \cite{cao2025uni3c}}                 & 83.55                                                             & 68.06                                                               & 53.16                                                                 & 66.09                                                              & 98.94                                                                 & 93.74                                                                      & 94.19                                                                   & 97.19                                                              & 97.05                                                           & 1.361           & 7.731             & 5.883          \\
\textbf{Ours}                  & \textbf{88.10}                                                    & \textbf{72.70}                                                      & \textbf{57.49}                                                        & \textbf{86.26}                                                     & 98.79                                                                 & \textbf{95.69}                                                             & \textbf{96.48}                                                          & \textbf{98.76}                                                     & \textbf{98.65}                                                  & \textbf{0.890}  & \textbf{3.103}    & \textbf{2.507} \\
    \bottomrule
\end{tabular}
\end{table*}

\noindent\textbf{Data annotation.}
For each retained clip, we automatically generate a text prompt and rendered 4D control maps.
We first generate a descriptive caption using Qwen2.5-VL-72B~\cite{bai2025qwen2}, which serves as text prompt.
For geometry, we adopt MegaSAM as the base pipeline and replace its monocular and metric depth modules with MoGe-2~\cite{wang2025moge} and UniDepth~V2~\cite{piccinelli2025unidepthv2}, respectively, to obtain more accurate and temporally consistent depth.
Given the video frames, the estimated depth maps, and the camera trajectory, we reconstruct a 3D point cloud for each frame.
Applying Grounded-SAM2 object masks to the per-frame point clouds yields per-object point clouds and a static background point cloud $P^{\text{bg}}$, as described in Sec.~\ref{sec:4d_control}.
For each object, we then fit per-frame 3D Gaussians and form a 3D Gaussian trajectory $\{\mathcal{G}_o^t\}$.
Finally, we render the 4D Geometric Control into model-ready 4D control maps.
The static background point cloud is rendered under the camera trajectory to obtain background RGB/depth maps.
The 3D Gaussian trajectories are rendered to obtain 3D Gaussian trajectories RGB/depth maps.
We then invert the background mask and merge it with the 3D Gaussian trajectories mask to produce a soft merged mask that marks regions where the video diffusion model should synthesize content.

In total, VerseControl4D contains 35{,}000 training samples and 1{,}000 validation samples. 
In the training set, about 26\% of samples are sourced from Sekai-Real-HQ and 74\% from SpatialVID-HQ, while 20\% of the samples depict static scenes, encouraging VerseCrafter to learn both camera-only world exploration and coupled camera–object dynamics. 
The validation set includes 250 static-scene samples to specifically assess camera-only control.

\section{Experiments}

\begin{figure}
    \centering
    \includegraphics[width=1\linewidth]{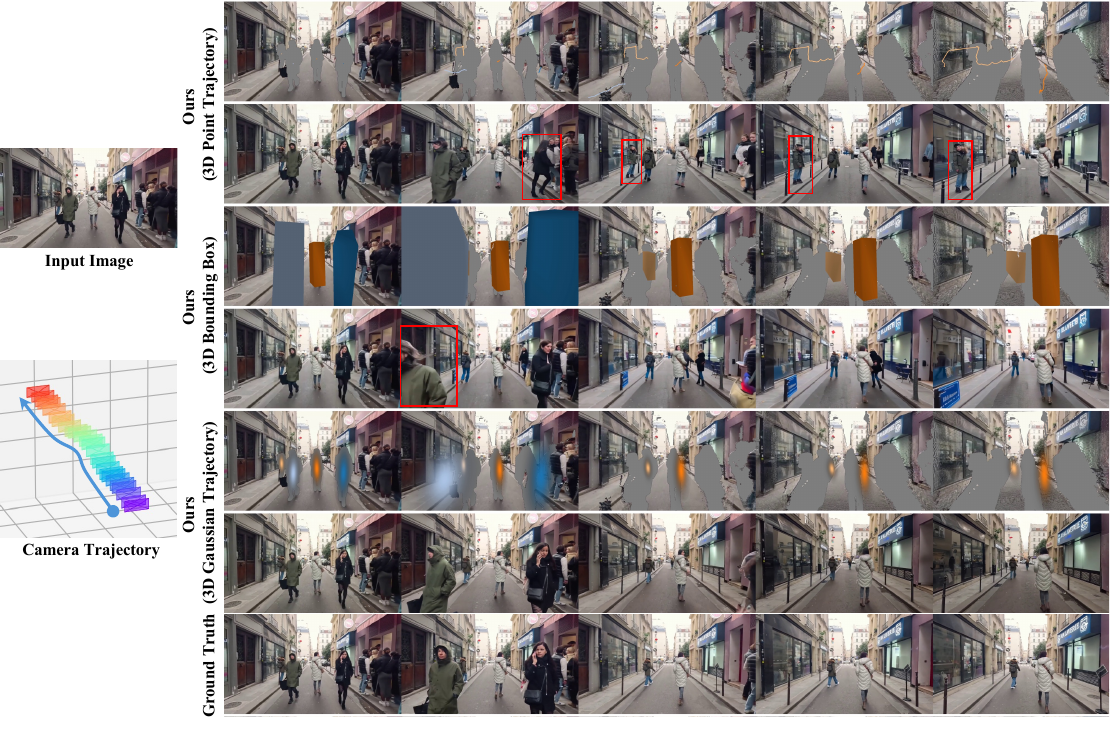}
    \caption{\textbf{Ablation on 3D representations for object motion control.}
    We compare object control using \emph{3D point trajectory} (top), \emph{3D bounding box} (middle), and \emph{3D Gaussian trajectory} (bottom). 
    3D point trajectory and 3D bounding box often cause scale drift and misaligned motion (red boxes), whereas 3D Gaussian trajectory better follows the intended object motion while preserving plausible shapes and background interactions. \label{fig:ablation_representation}}
\vspace{-5pt}
\end{figure}

\noindent\textbf{Implementation Details.}
We build VerseCrafter upon the Wan2.1 T2V-14B model. The Wan backbone is kept frozen, and only GeoAdapter is updated. Each GeoAdapter block is initialized from the weights of its paired DiT block in Wan-DiT to stabilize training, and we set $k=5$ so that every 5th DiT block in Wan-DiT is paired with a GeoAdapter block.
We use the Adam optimizer with a learning rate of 2e-5, a constant learning rate schedule with 100 warmup steps. All experiments are conducted on 16  96-GB GPUs with a global batch size of 16. Training is performed in two stages: we first train for 2{,}500 iterations on $480$P clips, and then fine-tune the same model for another 2{,}500 iterations on $720$P clips. The total wall-clock training time is about 380 hours.
We adopt classifier-free guidance during training by randomly dropping the text condition with probability 0.1. At inference time, we use 50 denoising steps and a classifier-free guidance scale of 5.0. Generating an 81-frame $720$P video clip on 8 96-GB GPU takes about 1152 seconds, with a peak per-GPU memory usage of about 90\,GB.

\setlength{\tabcolsep}{1pt}
\begin{table*}[]
\caption{\textbf{Camera-only motion control on static scenes.}
On the static subset of VerseControl4D, we report VBench-I2V scores and camera control metrics (RotErr, TransErr).
VerseCrafter achieves the best overall visual quality while substantially reducing camera pose errors.}
\label{tab:cam_only}
\scriptsize
\centering
\begin{tabular}{l|ccccccccc|cc}
\toprule
                     & \textbf{\begin{tabular}[c]{@{}c@{}}Overall \\ Score\end{tabular}$\uparrow$} & \textbf{\begin{tabular}[c]{@{}c@{}}Imaging \\ Quality\end{tabular}$\uparrow$} & \textbf{\begin{tabular}[c]{@{}c@{}}Aesthetic \\ Quality\end{tabular}$\uparrow$} & \textbf{\begin{tabular}[c]{@{}c@{}}Dynamic \\ Degree\end{tabular}$\uparrow$} & \textbf{\begin{tabular}[c]{@{}c@{}}Motion\\  Smoothness\end{tabular}$\uparrow$} & \textbf{\begin{tabular}[c]{@{}c@{}}Background \\ Consistency\end{tabular}$\uparrow$} & \textbf{\begin{tabular}[c]{@{}c@{}}Subject \\ Consistency\end{tabular}$\uparrow$} & \textbf{\begin{tabular}[c]{@{}c@{}}I2V\\  Background\end{tabular}$\uparrow$} & \textbf{\begin{tabular}[c]{@{}c@{}}I2V \\ Subject\end{tabular}$\uparrow$} & \textbf{RotErr$\downarrow$} & \textbf{TransErr$\downarrow$} \\
                     \midrule
\textbf{ViewCrafter \cite{yu2024viewcrafter}} & 84.04                                                             & 69.56                                                               & 55.52                                                                 & 68.02                                                              & 97.86                                                                 & 92.09                                                                      & 94.25                                                                   & 97.70                                                              & 97.29                                                           & 2.101           & 9.868             \\
\textbf{Voyager \cite{huang2025voyager}}     & 78.12                                                             & 55.48                                                               & 49.80                                                                 & 65.34                                                              & \textbf{99.39}                                                        & 92.31                                                                      & 91.55                                                                   & 86.02                                                              & 85.03                                                           & 3.557           & 3.880             \\
\textbf{FlashWorld \cite{li2025flashworld}}  & 85.33                                                             & 71.68                                                               & \textbf{58.74}                                                                 & 73.46                                                              & 98.35                                                                 & 94.27                                                                      &  92.47                                                                  &  95.38                                                             & 98.32                                                           & 1.792           & 3.257            \\
\textbf{Ours}        & \textbf{86.80}                                                    & \textbf{74.57}                                                      & 54.78                                                        & \textbf{80.34}                                                     & 97.62                                                                 & \textbf{94.88}                                                             & \textbf{95.55}                                                          & \textbf{97.86}                                                     & \textbf{98.79}                                                  & \textbf{0.650}  & \textbf{2.587}   \\
    \bottomrule
\end{tabular}
\end{table*}

\setlength{\tabcolsep}{1pt}
\begin{table*}[]
\caption{\textbf{Ablation study on 3D representation, depth, and decoupled controls.}
We compare different variants of VerseCrafter using VBench-I2V and 3D control metrics (RotErr, TransErr, ObjMC).
Our full model with 3D Gaussian trajectories, depth-aware rendering, and decoupled background/foreground controls achieves the best visual quality and the most accurate camera and object motion control.}
\label{tab:ablation}
\scriptsize
\centering
\begin{tabular}{l|ccccccccc|ccc}
\toprule
                                         & \textbf{\begin{tabular}[c]{@{}c@{}}Overall \\ Score\end{tabular}$\uparrow$} & \textbf{\begin{tabular}[c]{@{}c@{}}Imaging \\ Quality\end{tabular}$\uparrow$} & \textbf{\begin{tabular}[c]{@{}c@{}}Aesthetic \\ Quality\end{tabular}$\uparrow$} & \textbf{\begin{tabular}[c]{@{}c@{}}Dynamic \\ Degree\end{tabular}$\uparrow$} & \textbf{\begin{tabular}[c]{@{}c@{}}Motion \\ Smoothness\end{tabular}$\uparrow$} & \textbf{\begin{tabular}[c]{@{}c@{}}Background \\ Consistency\end{tabular}$\uparrow$} & \textbf{\begin{tabular}[c]{@{}c@{}}Subject \\ Consistency\end{tabular}$\uparrow$} & \textbf{\begin{tabular}[c]{@{}c@{}}I2V \\ Background\end{tabular}$\uparrow$} & \textbf{\begin{tabular}[c]{@{}c@{}}I2V \\ Subject\end{tabular}$\uparrow$} & \textbf{RotErr$\downarrow$} & \textbf{TransErr$\downarrow$} & \textbf{ObjMC$\downarrow$} \\
                                         \midrule
\textbf{Ours (3D Bounding Box)}          & 85.45                                                             & 69.23                                                               & 55.70                                                                 & 78.57                                                              & 98.70                                                                 & 92.92                                                                      & 93.27                                                                   & 97.74                                                              & 97.48                                                           & 1.350           & 3.805             & 4.520          \\
\textbf{Ours (3D Point Trajectory)}      & 85.57                                                             & 70.29                                                               & 55.27                                                                 & 78.23                                                              & 98.63                                                                 & 94.00                                                                      & 92.75                                                                   & 97.85                                                              & 97.55                                                           & 1.298           & 3.281             & 6.896          \\
\textbf{Ours (w/o depth)}                & 85.64                                                             & 70.19                                                               & 55.00                                                                 & 80.60                                                              & 98.66                                                                 & 92.07                                                                      & 92.83                                                                   & 98.07                                                              & 97.69                                                           & 1.177           & 3.900             & 4.929          \\
\textbf{Ours (BG \& FG Merged)} & 85.72                                                             & 69.19                                                               & 54.86                                                                 & 83.72                                                              & 98.65                                                                 & 91.15                                                                      & 92.86                                                                   & 97.93                                                              & 97.41                                                           & 1.080           & 3.803             & 3.726          \\
\textbf{Ours}                            & \textbf{88.10}                                                    & \textbf{72.70}                                                      & \textbf{57.49}                                                        & \textbf{86.26}                                                     & \textbf{98.79}                                                        & \textbf{95.69}                                                             & \textbf{96.48}                                                          & \textbf{98.76}                                                     & \textbf{98.65}                                                  & \textbf{0.890}           & \textbf{3.103}    & \textbf{2.507}      \\
\bottomrule
\end{tabular}
\end{table*}

\noindent\textbf{Evaluation Metrics.}
We evaluate overall video quality using VBench-I2V.
For camera control, we follow CameraCtrl~\cite{he2024cameractrl} and report rotation error (RotErr) and translation error (TransErr).
For object-motion control, we adopt ObjMC proposed in MotionCtrl~\cite{wang2024motionctrl}.
Given a generated video, we apply the same geometry annotation pipeline used for VerseControl4D to estimate its camera trajectory and 3D Gaussian trajectories, and compare them with the corresponding ground-truth trajectories from our dataset.
ObjMC is computed as the average Euclidean distance between the estimated and ground-truth 3D Gaussian means over all controlled objects and frames.

\subsection{Joint Camera and Object Motion Control}
We first evaluate joint control of camera and object motion on VerseControl4D. 
As shown in Table~\ref{tab:joint_control}, VerseCrafter achieves the best VBench-I2V scores among all compared methods, with clear gains in Overall Score, Imaging Quality, Aesthetic Quality, and both subject  and background consistency.
On 3D control metrics, VerseCrafter substantially reduces rotation, translation, and object-motion errors compared with the best-performing baseline, reflecting much tighter alignment with the target 4D geometric control.
Qualitative comparisons in Fig.~\ref{fig:joint_control_vis} further highlight these differences: Perception-as-Control and Uni3C exhibit noticeable human deformation, while Yume roughly follows the text-described motion but lacks precise camera control. 
Uni3C, relying on SMPL-X, is limited to single-person motion and struggles with other categories such as vehicles.
In contrast, VerseCrafter more faithfully follows both the camera trajectory and 3D Gaussian trajectories while maintaining sharp appearance and geometrically consistent backgrounds.

\begin{figure}
    \centering
    \includegraphics[width=1\linewidth]{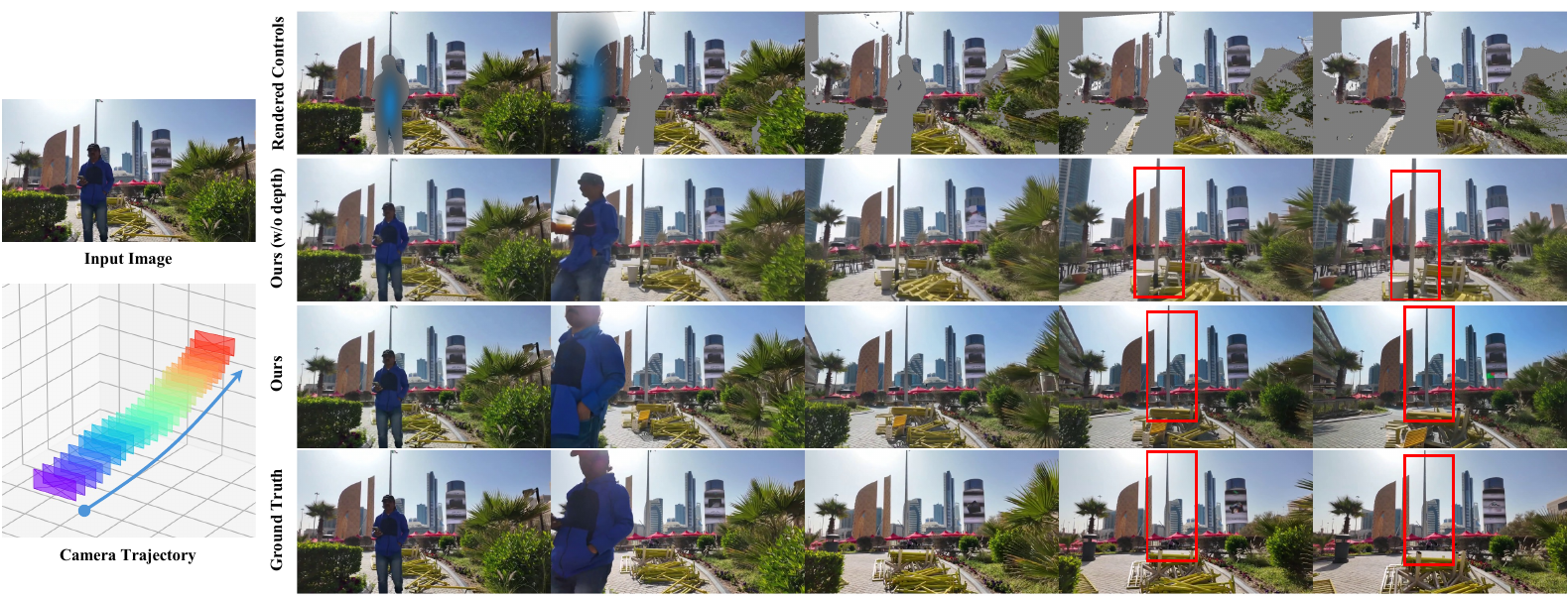}
    \caption{\textbf{Ablation on depth-aware control.}
    We compare VerseCrafter without depth inputs (\emph{Ours (w/o depth)}, top) and with RGB+depth inputs (middle) under the same camera trajectory.
    Without depth, the model often produces incorrect foreground-background ordering, e.g., lampposts are pulled in front of distant buildings, and occlusion boundaries drift over time (red boxes).
    With RGB+depth, the model recovers consistent parallax and occlusion, producing geometry much closer to the ground truth.
\label{fig:ablation_depth}}
\vspace{-5pt}
\end{figure}

\subsection{Camera-Only Motion Control}

We evaluate camera-only control on the static-scene subset of VerseControl4D, where objects remain stationary and only the camera moves.
As shown in Table~\ref{tab:cam_only}, VerseCrafter achieves the best VBench-I2V performance among all compared methods, with consistent gains in Overall Score, Imaging Quality, and both subject and background consistency, while maintaining motion smoothness comparable to prior methods.
On 3D camera metrics, VerseCrafter substantially reduces rotation and translation errors relative to the best-performing baseline, indicating that it follows the target camera trajectory much more faithfully in static scenes.
Qualitative comparisons in Fig.~\ref{fig:camera_only_vis} further confirm these trends: ViewCrafter and Voyager exhibit distorted facades, drifting structures, or inaccurate camera motion, while FlashWorld tends to produce blurred scene boundaries and imprecise camera motion. In contrast, VerseCrafter preserves straight structures, stable depth relationships, and an appearance closer to the ground-truth video, evidencing precise camera control in a static scene.

\begin{figure}
    \centering
    \includegraphics[width=1\linewidth]{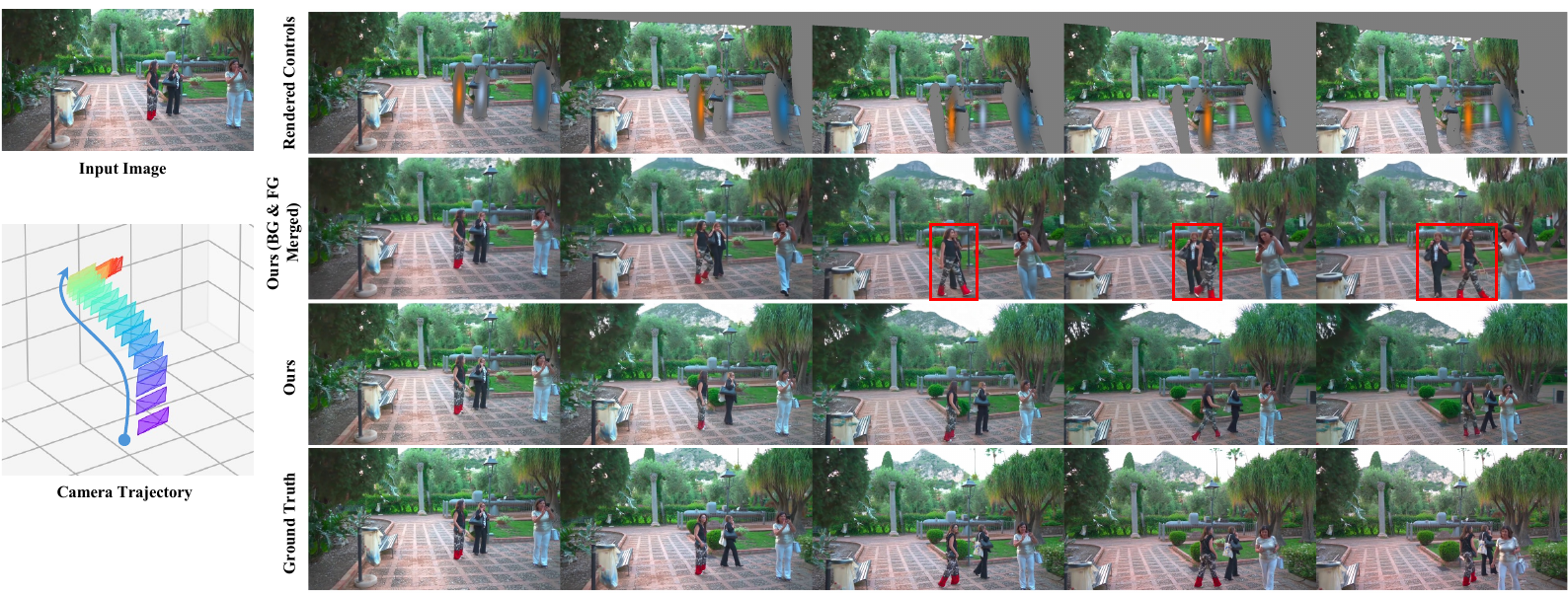}
    \caption{\textbf{Ablation on decoupled background and foreground controls.}
    We compare a variant that merges background and foreground controls into a single map (\emph{Ours (BG \& FG Merged)}, top) with our default decoupled design (middle).
    When the controls are merged, object motion control degrades significantly (red boxes), whereas the decoupled design better preserves the static background and produces more accurate and stable object motion.
    \label{fig:ablation_merged}}
\end{figure}

\subsection{Ablation Study}

We conduct ablations to analyze three key design choices in VerseCrafter:
(i) the 3D representations for object motion,
(ii) the use of depth in control maps, and
(iii) the decoupling of background and foreground controls.
All variants share the same training data, backbone, and optimization settings; only the control design is changed.

\noindent \textbf{3D representations for object motion.}
To isolate the effect of our motion representation, we derive two alternatives from per-object 3D Gaussian trajectories:
(1) an oriented \textbf{3D bounding box} whose axes follow the Gaussian's principal directions and whose side lengths are scaled by its principal spreads; and
(2) a \textbf{3D point trajectory} that retains only the Gaussian centroid.
The rest of the pipeline is unchanged: we rasterize cuboids (for boxes) or tiny disks/spheres (for points) instead of Gaussian ellipses.
As reported in Table~\ref{tab:ablation}, replacing Gaussians with boxes slightly hurts both visual quality and control accuracy, while point trajectories give the weakest object-motion consistency.
Qualitatively (Fig.~\ref{fig:ablation_representation}), points and boxes often cause scale drift and misaligned motion, whereas 3D Gaussian trajectories better follow the intended object trajectories and preserve plausible object shapes.

\noindent \textbf{Depth-aware Control.}
To evaluate the effect of depth, we remove depth channel from the background and trajectory controls (``Ours (w/o depth)'' in Table~\ref{tab:ablation}). This variant yields a lower Overall Score and significantly worse 3D control (higher RotErr and ObjMC values). As shown in Fig.~\ref{fig:ablation_depth}, without depth, the model produces incorrect foreground-background ordering: vertical structures like streetlights appear beside shelves in the foreground, while buildings that should be behind the character are positioned elsewhere, and occlusion boundaries drift over time. With RGB+depth, VerseCrafter recovers more consistent parallax and occlusion, producing geometry much closer to the ground truth.

\noindent\textbf{Decoupled vs.\ merged controls.}
We further compare our decoupled design with a variant that merges the background and 3D Gaussian trajectory maps into a single control stream (\emph{Ours (BG \& FG Merged)} in Table~\ref{tab:ablation}).
Although this variant still benefits from the explicit 4D geometric scene state, it consistently underperforms the full model, with a particularly noticeable drop in object-motion accuracy.
As shown in Fig.~\ref{fig:ablation_merged}, merging the controls leads to clear degradation in object motion control.
In contrast, keeping decoupled design preserves static geometry while producing more precise and stable object motion, which is crucial for accurate and geometry-consistent control.

\section{Conclusion}

We present \textbf{VerseCrafter}, a geometry-driven video world model built upon an explicit \textbf{4D Geometric Control}, represented by a static background point cloud and per-object 3D Gaussian trajectories in a shared world coordinate frame.
Coupled with GeoAdapter, which conditions a frozen Wan2.1 backbone on rendered 4D control maps, this design enables high-fidelity video generation with precise, disentangled control over camera and multi-object motion. To support training and evaluation, we construct \textbf{VerseControl4D}, a real-world dataset with automatically derived prompts and rendered 4D control maps, comprising 35K training samples. Experiments and ablations show that VerseCrafter delivers superior visual quality and more accurate joint camera and object motion than existing controllable video generators and video world models, highlighting 4D Geometric Control as a promising interface for future work on dynamic world simulation and editing.

\paragraph*{Acknowledgments.}

The paper is supported by Shanghai Municipal Science Technology Major Project (2025SHZDZX025G02).

% \clearpage
% \setcounter{page}{1}
% \maketitlesupplementary

\appendix

\begin{figure*}[t]
    \centering
    \includegraphics[width=0.9\linewidth]{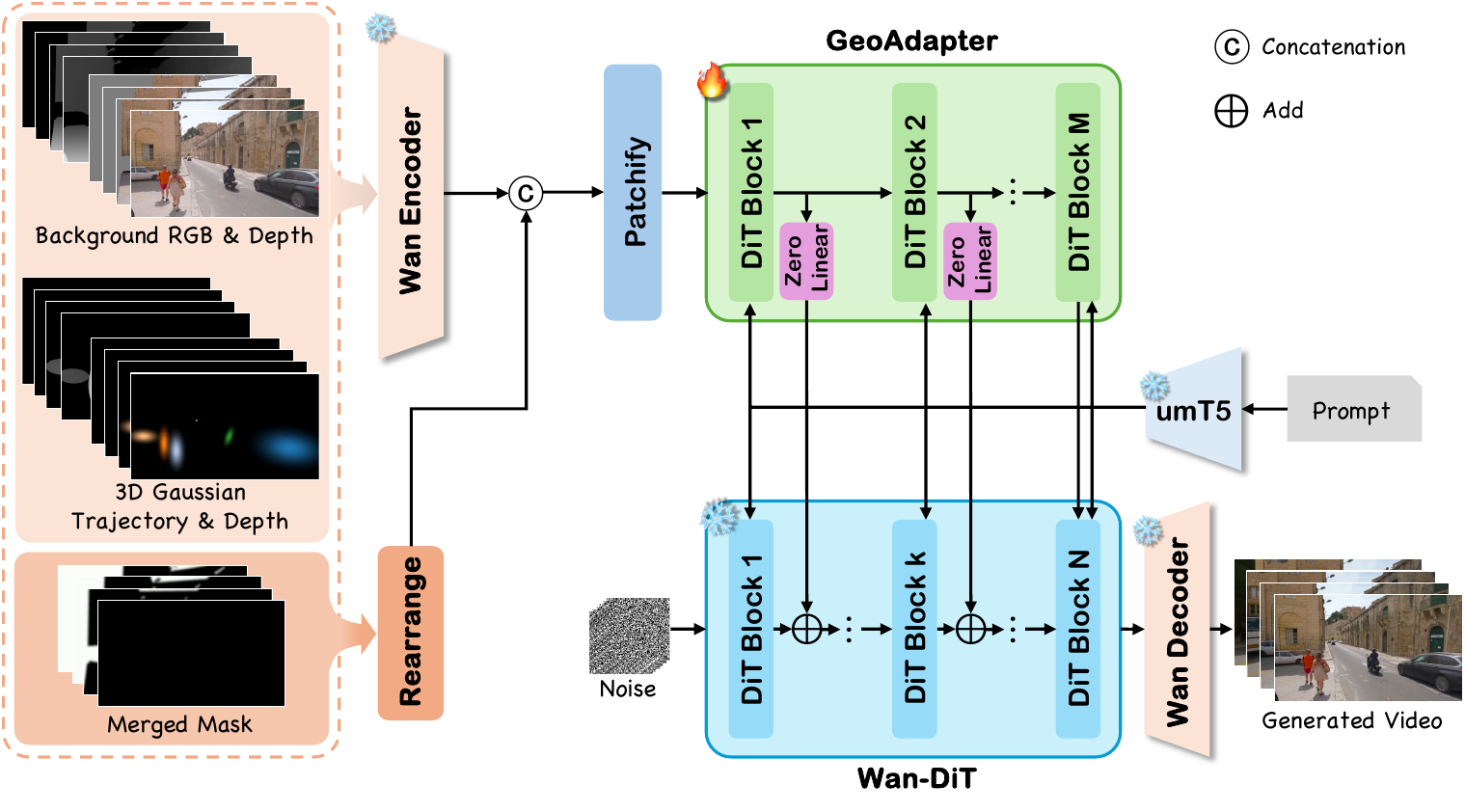}
\caption{\textbf{Detailed architecture of VerseCrafter.}
Background RGB \& depth maps and 3D Gaussian trajectory RGB \& depth maps are first encoded by the frozen Wan Encoder.
The soft merged mask is rearranged into latent-aligned channels, and all geometry latents are then concatenated along the channel dimension to form a unified spatio-temporal geometry feature.
This feature is patchified into tokens and processed by the GeoAdapter branch.
At selected Wan-DiT blocks, GeoAdapter outputs are passed through zero-initialized linear layers and added to the backbone tokens as residual modulations, enabling geometry-consistent control over camera motion and multi-object motion.}
    \label{fig:detailed_arch}
\end{figure*}

\section{Preliminary: Video Diffusion Models}

Modern video diffusion models operate in a compact latent space learned by a spatio-temporal VAE. Given a video $x \in \mathbb{R}^{T \times H \times W \times 3}$, the encoder $E$ maps it to latent features $z_0 = E(x) \in \mathbb{R}^{T' \times C \times H' \times W'}$, on which the generative process is defined~\cite{rombach2022high,blattmann2023align}. A standard forward diffusion process gradually perturbs $z_0$ into noisy variables $z_t$ via
\begin{equation}
q(z_t \mid z_0) = \sqrt{\alpha_t}\, z_0 + \sqrt{1-\alpha_t}\, \epsilon, \quad \epsilon \sim \mathcal{N}(0, I),
\end{equation}
and a denoiser $\epsilon_\theta$ is trained to predict the noise under time step $t$ and conditioning signal $c$ (e.g., text prompts or reference frames) as
\begin{equation}
\mathcal{L}_{\text{diff}}(\theta) = \mathbb{E}_{z_0, t, \epsilon} \big[ \| \epsilon_\theta(z_t, t, c) - \epsilon \|_2^2 \big],
\end{equation}
following the DDPM formulation~\cite{ho2020denoising}. Recent video generators further adopt continuous-time flow matching. Given clean latents $z_0$ and a Gaussian samples $z_1$, one defines linear interpolants $z_\tau = (1-\tau) z_0 + \tau z_1$ with $\tau \in [0,1]$ and learns a velocity field $v_\theta$ by
\begin{equation}
\mathcal{L}_{\text{flow}}(\theta) = \mathbb{E}_{z_0, \tau, \epsilon} \big[ \| v_\theta(z_\tau, \tau, c) - (z_1 - z_0) \|_2^2 \big],
\end{equation}
as in flow-matching and related ODE-based generative formulations~\cite{lipman2022flow,karras2022elucidating}. These objectives are typically implemented with Diffusion Transformers (DiT), which operate on spatio-temporal latent tokens and inject $(t, c)$ through attention~\cite{peebles2023scalable}, forming the backbone of current foundation video generators.

Wan2.1 instantiates the above latent video diffusion / flow-matching paradigm with a 3D VAE and a DiT-based denoiser, together with rich multi-modal conditioning trained on large-scale, diverse video--text data~\cite{wan2025wan}. In VerseCrafter, we adopt Wan2.1-14B as a frozen latent video diffusion backbone and treat it as a generic video prior. Specifically, we keep the Wan Encoder, Wan-DiT, and Wan Decoder unchanged, and  attach a lightweight geometry-aware control interface, namely GeoAdapter, to selected Wan-DiT blocks. The detailed architecture of VerseCrafter is provided in the Sec. \ref{sec:supp_arch}.

\section{Model Architecture Details}
\label{sec:supp_arch}

VerseCrafter is built on top of the Wan2.1 T2V-14B backbone~\cite{wan2025wan}, a latent video diffusion / flow-matching model with a 3D VAE (Wan Encoder and Wan Decoder) and a DiT-based denoiser (Wan-DiT).
As shown in Fig.~\ref{fig:detailed_arch}, we keep the Wan2.1 backbone frozen and introduce a geometry-aware conditioning pathway with a lightweight GeoAdapter that conditions selected Wan-DiT blocks on rendered 4D control maps. 
Table~\ref{tab:model_variants} summarizes the input resolution, number of Wan-DiT layers, hidden dimension, GeoAdapter injection pattern, and fine-tuning configuration of VerseCrafter.

\setlength{\tabcolsep}{1pt}
\begin{table}[t]
\centering
\caption{\textbf{Model configuration of VerseCrafter.}
Settings include the final output resolution, number of Wan-DiT layers, GeoAdapter injection blocks, pre-trained backbone, and training schedule.}
\label{tab:model_variants}
\small
\begin{tabular}{l|c}
\toprule
 & \textbf{VerseCrafter} \\
\midrule
Final resolution & $720$P \\
Num. Wan-DiT layers & 40 \\
GeoAdapter injection blocks & $[0, 5, 10, 15, 20, 25, 30, 35]$ \\
Pre-trained backbone & Wan2.1 T2V-14B \\
Hidden dimension & 5120 \\
Batch size & 16 \\
Training schedule & 2{,}500 it. @480P + 2{,}500 it. @720P \\
\bottomrule
\end{tabular}
\end{table}

\noindent\textbf{Geometry encoding and tokenization.}
For each frame $t$, we render background RGB/depth $\mathrm{RGB}^{\mathrm{bg}}_t$, $\mathrm{Depth}^{\mathrm{bg}}_t$, 3D Gaussian trajectory RGB/depth $\mathrm{RGB}^{\mathrm{traj}}_t$, $\mathrm{Depth}^{\mathrm{traj}}_t$, and a soft merged mask $M_t$ that marks regions where the diffusion model should synthesize or overwrite content. For $t{=}1$, we replace $\mathrm{RGB}^{\mathrm{bg}}_1$ with the input image and set $M_1{=}0$.
The four RGB/depth maps are encoded by the frozen Wan Encoder to obtain latent features at the VAE resolution, while the mask $M \in \mathbb{R}^{1 \times T \times H \times W}$ is rearranged to align with the latent grid of Wan Encoder (the ``Rearrange'' module in Fig.~\ref{fig:detailed_arch}).
Let $s_t$, $s_h$, and $s_w$ denote the temporal and spatial strides of Wan Encoder (we use $s_t{=}4$ and $s_h{=}s_w{=}8$).
Following the practice in~\cite{jiang2025vace,wan2025wan}, we drop the singleton channel dimension, split the spatial dimensions into $s_h \times s_w$ sub-cells, and fold these sub-cells into the channel dimension via a reshape--permute operation, yielding a tensor of shape $C_M \times T \times H' \times W'$ with $C_M{=}s_h s_w$, $H'{=}H/s_h$, and $W'{=}W/s_w$.
We then downsample the temporal dimension using nearest-neighbor interpolation to match the latent depth $T' = (T + s_t - 1)/s_t$, producing $\hat{M} \in \mathbb{R}^{C_M \times T' \times H' \times W'}$.
Finally, $\hat{M}$ is concatenated channel-wise with the encoded background and 3D Gaussian trajectory latents to form a unified spatio-temporal geometry feature
$\mathcal{G} \in \mathbb{R}^{T' \times H' \times W' \times C_{\mathcal{G}}}$.
Following Wan-DiT, we partition $\mathcal{G}$ into non-overlapping 3D patches and linearly project each patch into a token embedding, yielding a sequence of geometry tokens
$\mathbf{g} \in \mathbb{R}^{L \times D}$,
where $L = T' H' W'$ is the number of spatio-temporal patches and $D$ matches the hidden width of Wan-DiT.
Because we use identical strides, positional encodings, and patch sizes, the geometry tokens are spatially and temporally aligned with the latent video tokens processed by Wan-DiT.

\noindent\textbf{GeoAdapter integration.}
GeoAdapter is a lightweight DiT-style branch  that operates on geometry tokens  $\mathbf{g}$.
It shares the same hidden dimensionality and positional encodings as Wan-DiT, but contains far fewer layers.
Let $\{\mathcal{B}_1,\dots,\mathcal{B}_{N}\}$ denote $N$ Wan-DiT blocks, and let $\{\mathcal{G}_1,\dots,\mathcal{G}_M\}$ denote $M$ GeoAdapter blocks.
We attach GeoAdapter as a residual modulation branch to a subset of Wan-DiT blocks.
Concretely, we choose a stride $k$ and inject GeoAdapter after every $k$-th Wan-DiT block; see Table~\ref{tab:model_variants} for the exact injection pattern and configuration.
For each Wan-DiT block $\mathcal{B}_n$ whose index $n$ belongs to the injection set, with input tokens $\mathbf{x}_n \in \mathbb{R}^{L \times D}$ and geometry tokens $\mathbf{g}$, we add a geometry-conditioned residual of the form
\begin{equation}
\mathbf{x}_{n+1}
= \mathcal{B}_n(\mathbf{x}_n)
+ \mathcal{G}_m(\mathbf{g}) \,\mathbf{W}^{(m)}_0,
\end{equation}
where $\mathcal{G}_m$ is the corresponding GeoAdapter block and $\mathbf{W}^{(m)}_0 \in \mathbb{R}^{D \times D}$ is its output projection.
Each GeoAdapter block is initialized from the weights of its paired Wan-DiT block for stable training, while $\mathbf{W}^{(m)}_0$ is initialized to zero.
As a result, VerseCrafter behaves identically to the original Wan2.1 backbone at the beginning of training.
During fine-tuning, $\mathbf{W}^{(m)}_0$ gradually learns to inject geometry information through residual modulation, in the spirit of zero-initialized adapter designs such as ControlNet~\cite{zhang2023adding}.

\setlength{\tabcolsep}{3pt}
\begin{table}[t]
\centering
\caption{\textbf{VerseControl4D data split and scene-type statistics.}
We report the number of samples from each source dataset and split.
\emph{Dynamic scenes} contain coupled camera motion and foreground object motion,
while \emph{static scenes} have negligible object motion and are used for camera-only evaluation.}
\label{tab:versecontrol4d_stats}
\small
\begin{tabular}{lccc}
\toprule
\multirow{2}{*}{\textbf{Split}} 
& \textbf{Sekai-Real-HQ}
& \multicolumn{2}{c}{\textbf{SpatialVID-HQ}} \\
\cmidrule(lr){2-2}\cmidrule(lr){3-4}
& \textbf{Dynamic Scenes} & \textbf{Static Scenes} & \textbf{Dynamic Scenes} \\
\midrule
Train      & 9{,}071 & 7{,}000 & 18{,}929 \\
Validation & 468     & 250     & 282 \\
\bottomrule
\end{tabular}
\end{table}

\begin{figure*}[t]
    \centering
    \includegraphics[width=0.9\linewidth]{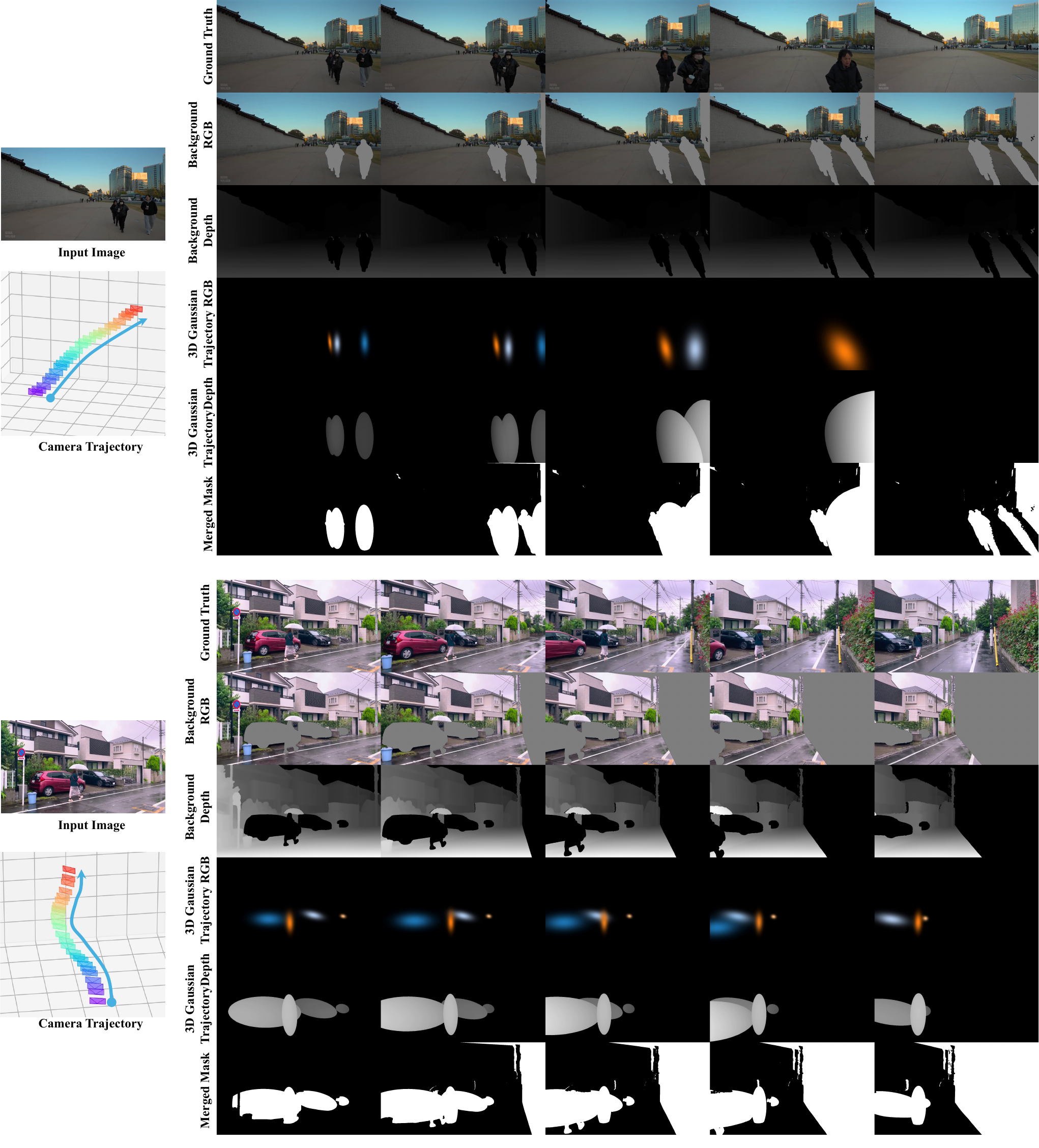}
    \caption{\textbf{VerseControl4D dataset examples.}
    For each clip, we visualize the input image and target camera trajectory (left),
    followed by several frames of ground-truth video and our rendered control signals (right):
    background RGB/depth, 3D Gaussian trajectory RGB/depth for controlled objects, and the final merged mask.
    These signals are automatically derived by our annotation pipeline in main paper.}
    \label{fig:dataset_examples}
\end{figure*}

\section{VerseControl4D Dataset Details}
\label{sec:versecontrol4d}

We construct \textbf{VerseControl4D}, a large-scale real-world video dataset with automatically derived prompts and rendered 4D control maps.
As described in the main paper, VerseControl4D is built through four stages: data collection, clip extraction, quality filtering, and data annotation.
The rendered 4D control maps comprise background RGB/depth maps, 3D Gaussian trajectory RGB/depth maps, and a soft merged mask.

VerseControl4D contains 35{,}000 training samples and 1{,}000 validation samples.
Table~\ref{tab:versecontrol4d_stats} summarizes the data distribution by source dataset and scene type. Overall, 26\% of the samples come from Sekai-Real-HQ and 74\% from SpatialVID-HQ, reflecting their complementary scene coverage.
To support both camera-only world exploration and joint camera-object control, VerseControl4D includes
\emph{dynamic scenes} (clips with salient foreground object motion together with camera motion) and
\emph{static scenes} (clips with negligible object motion and only camera motion).
About 20\% of the training samples are \emph{static scenes}, and the validation set includes 250 static-scene samples for dedicated camera-only evaluation.
Representative samples and their rendered 4D control signals are shown in Fig.~\ref{fig:dataset_examples}.

\section{Evaluation Metrics}

\subsection{VBench-I2V}

We evaluate image-to-video generation quality using the VBench Image-to-Video (I2V) evaluation suite, denoted as VBench-I2V.
For each generated clip, we follow the official VBench-I2V protocol: the conditioning image and its corresponding generated video are fed into the evaluation pipeline, which computes a set of learned, human-aligned metrics that jointly capture video-image consistency and perceptual video quality.
In our experiments, we report the following eight VBench-I2V dimensions, and define the \emph{Overall Score} as the simple arithmetic mean of these eight normalized scores, where higher values indicate better performance:

\begin{itemize}
  \item \textbf{Imaging Quality.}
  This metric measures low-level image fidelity, including sharpness and the absence of artifacts such as blur, noise, or overexposure. 
  VBench uses an image quality predictor (e.g., MUSIQ) and averages its scores across frames to obtain a video-level imaging-quality score. 

  \item \textbf{Aesthetic Quality.}
  This dimension assesses the artistic and aesthetic appeal of individual frames, including composition, color harmony, and realism.
  VBench applies an aesthetic quality predictor (e.g., the LAION aesthetic model) to each frame and averages the predictions over the clip.

  \item \textbf{Dynamic Degree.}
  This metric quantifies how dynamic the generated video is.
  Optical flow magnitudes (e.g., estimated by RAFT) are used to measure the amount of motion, and the score reflects whether the model produces sufficiently active (non-static) content.

  \item \textbf{Motion Smoothness.}
  This metric evaluates whether subject and camera motion evolve smoothly and follows reasonable physical dynamics.
  VBench leverages a pre-trained video frame interpolation prior to assess how well intermediate motion can be interpolated, with smoother and more physically plausible motion receiving higher scores.

  \item \textbf{Background Consistency.}
  This dimension measures the temporal stability of background layout and texture.
  Frame-level features (e.g., CLIP) are compared across time; large feature variations indicate flickering or unstable backgrounds and lead to lower scores.

  \item \textbf{Subject Consistency.}
  This dimension evaluates the temporal consistency of the foreground subject \emph{within} the video, regardless of the input image.
  VBench computes subject-region features across frames and measures their similarity over time to penalize identity drift or sudden appearance changes.

  \item \textbf{I2V Background (Video--Image Background Consistency).}
  This metric evaluates how well the global background in the video matches the background in the input image, especially for scene-centric inputs.
  VBench uses background-sensitive features (e.g., DreamSim) and aggregates image--frame and inter-frame similarities into a single background consistency score.

  \item \textbf{I2V Subject (Video--Image Subject Consistency).}
  This metric measures how well the main subject in the generated video matches the subject in the input image.
  VBench extracts high-level visual features (e.g., DINO) from the conditioning image and from each video frame, and combines image--frame similarities with inter-frame similarities into a weighted average subject consistency score.

\end{itemize}

Formally, given these eight per-dimension scores $\{s_k\}_{k=1}^8$ returned by VBench-I2V for a video, we define
\begin{equation}
\text{Overall Score} = \frac{1}{8} \sum_{k=1}^{8} s_k,
\end{equation}
which is the value reported as ``Overall Score'' in the main paper.

\subsection{Rotation Error (RotErr)}
To measure how well the generated camera motion follows the ground-truth camera trajectory, we adopt the camera-alignment metric from CameraCtrl~\cite{he2024cameractrl}.
For each generated video, we estimate its camera trajectory using the same geometry-annotation pipeline used for VerseControl4D, yielding rotation matrices $\{\mathbf{R}^{j}_{\mathrm{gen}}\}_{j=1}^{n}$ and translation vectors $\{\mathbf{T}^{j}_{\mathrm{gen}}\}_{j=1}^{n}$, where $n$ is the number of frames.
Let $\{\mathbf{R}^{j}_{\mathrm{gt}}\}_{j=1}^{n}$ denote the corresponding ground-truth rotation matrices.
The rotation error is computed by comparing the ground-truth and generated rotation matrices at each frame:
\begin{equation}
\mathrm{RotErr}
= \sum_{j=1}^{n}
\arccos\!\left(
\frac{\operatorname{tr}\!\big(\mathbf{R}^{j}_{\mathrm{gen}} {\mathbf{R}^{j}_{\mathrm{gt}}}^{\top}\big) - 1}{2}
\right),
\label{eq:roterr}
\end{equation}
where $\operatorname{tr}(\cdot)$ denotes the matrix trace.
Lower RotErr indicates better alignment between the generated and ground-truth camera orientations.

\subsection{Translation Error (TransErr)}
We also evaluate the accuracy of the generated camera positions.
Let $\{\mathbf{T}^{j}_{\mathrm{gt}}\}_{j=1}^{n}$ and $\{\mathbf{T}^{j}_{\mathrm{gen}}\}_{j=1}^{n}$ be the ground-truth and generated camera translation vectors for a video with $n$ frames.
Following CameraCtrl~\cite{he2024cameractrl}, the translation error is defined as the sum of per-frame Euclidean distances between the translation vectors:
\begin{equation}
\mathrm{TransErr}
= \sum_{j=1}^{n}
\bigl\| \mathbf{T}^{j}_{\mathrm{gt}} - \mathbf{T}^{j}_{\mathrm{gen}} \bigr\|_{2},
\label{eq:transerr}
\end{equation}
Lower TransErr indicates that the generated camera positions more closely match the ground-truth camera positions.

\subsection{Object Motion Control (ObjMC)}

For object-motion control, we follow the ObjMC metric proposed in MotionCtrl~\cite{wang2024motionctrl} and extend it to the multi-object setting under our 3D Gaussian trajectory representation.
Given a generated video, we run the same geometry annotation pipeline used for VerseControl4D to estimate per-object 3D Gaussian trajectories, and compare them with the corresponding ground-truth trajectories from our dataset.

Let $N_{\text{gt}}$ and $N_{\text{pred}}$ denote the numbers of ground-truth and predicted controlled objects in a sample, and let $T$ be the number of frames.
For each ground-truth object $o \in \{1,\dots,N_{\text{gt}}\}$ and frame $t \in \{1,\dots,T\}$, we denote the ground-truth 3D Gaussian mean by $\boldsymbol{\mu}^{(t)}_o \in \mathbb{R}^3$ and the estimated mean from the generated video by $\hat{\boldsymbol{\mu}}^{(t)}_k \in \mathbb{R}^3$ for a predicted object $k$.

\paragraph{Multi-object matching.}
Since $N_{\text{gt}}$ and $N_{\text{pred}}$ may differ, we first define the trajectory distance between a ground-truth object $o$ and a predicted object $k$ as the average Euclidean distance between their 3D Gaussian means over time:
\begin{equation}
d(o,k) = \frac{1}{T} \sum_{t=1}^{T}
\bigl\|\hat{\boldsymbol{\mu}}^{(t)}_k - \boldsymbol{\mu}^{(t)}_o\bigr\|_2.
\label{eq:traj_dist}
\end{equation}
We then build a cost matrix $\mathbf{C} \in \mathbb{R}^{N_{\text{gt}} \times N_{\text{pred}}}$ with entries $C_{ok} = d(o,k)$.
To handle unmatched objects, we pad this matrix with dummy rows and columns and fill them with a constant penalty $\lambda$ (set to $10.0$\,m in our experiments).
Finally, we apply the Hungarian algorithm~\cite{kuhn1955hungarian} to obtain an optimal one-to-one matching between ground-truth and predicted trajectories.
This step assigns each ground-truth object either to a predicted trajectory or to a dummy entry when no suitable match exists.

\paragraph{ObjMC score.}
Given the optimal matching, we define the per-object trajectory error for a ground-truth object $o$ as
\begin{equation}
d_o =
\begin{cases}
d(o,k) & \text{if } o \text{ is matched to a predicted object } k, \\
\lambda & \text{if } o \text{ is unmatched},
\end{cases}
\end{equation}
and compute the final ObjMC score as the average over all ground-truth controlled objects:
\begin{equation}
\text{ObjMC} = \frac{1}{N_{\text{gt}}} \sum_{o=1}^{N_{\text{gt}}} d_o.
\end{equation}
Lower ObjMC indicates more accurate multi-object 3D motion control, and the unmatched penalty $\lambda$ penalizes missed objects under the one-to-one matching formulation.

\section{Additional Qualitative Results}
\label{sec:supp_results}

\begin{figure*}
    \centering
    \includegraphics[width=0.9\linewidth]{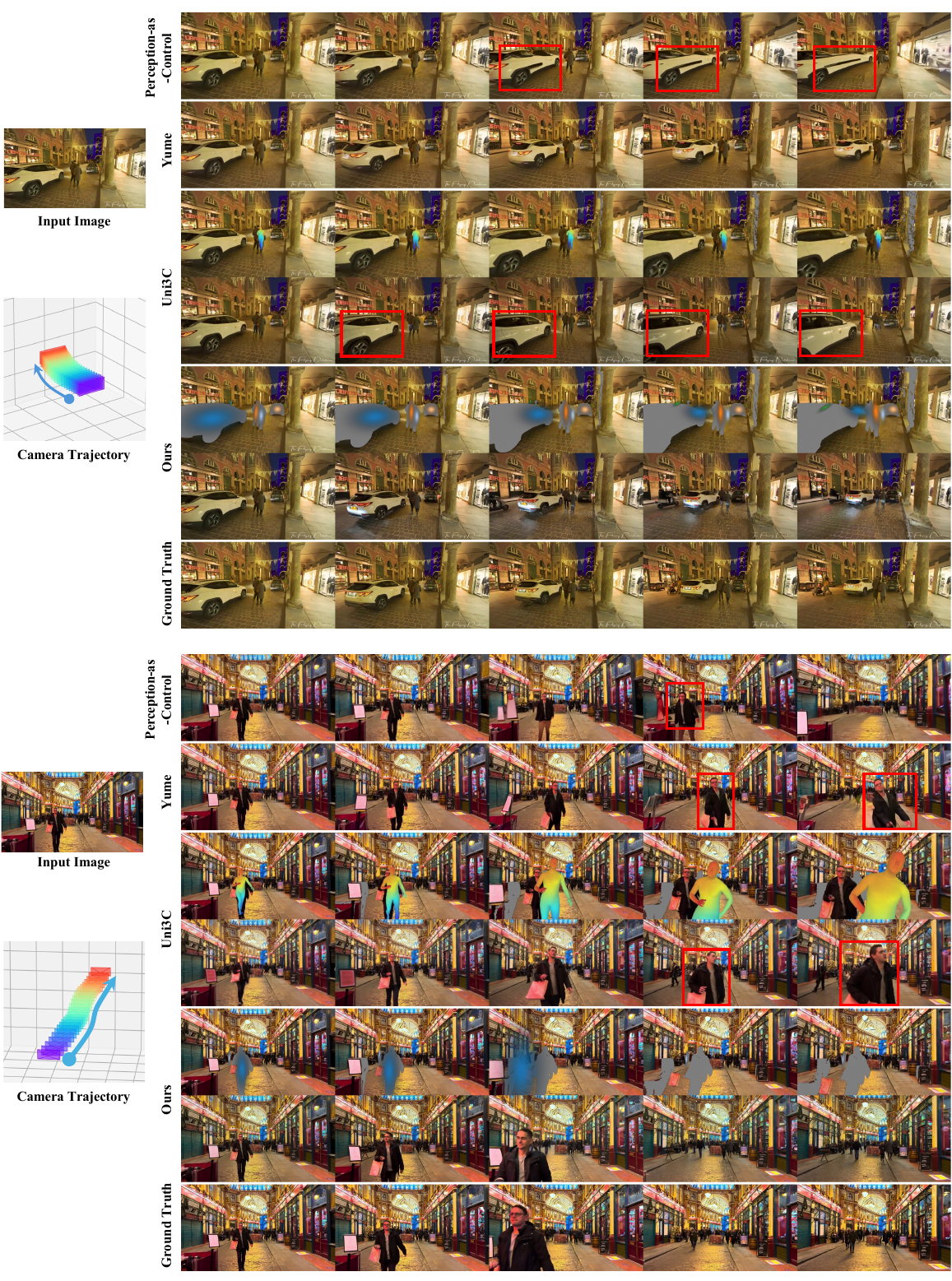}
\caption{\textbf{Additional qualitative comparison of joint camera and object motion control.}
Perception-as-Control and Uni3C exhibit noticeable object deformation, while Yume roughly follows the text-described motion but lacks precise camera control. Uni3C is also limited to a single human. In contrast, VerseCrafter more faithfully follows both the camera trajectory and multi-object motion while maintaining sharp appearance and geometrically consistent backgrounds.
\label{fig:supp_joint_control_vis1}}
\end{figure*}

\begin{figure*}
    \centering
    \includegraphics[width=0.9\linewidth]{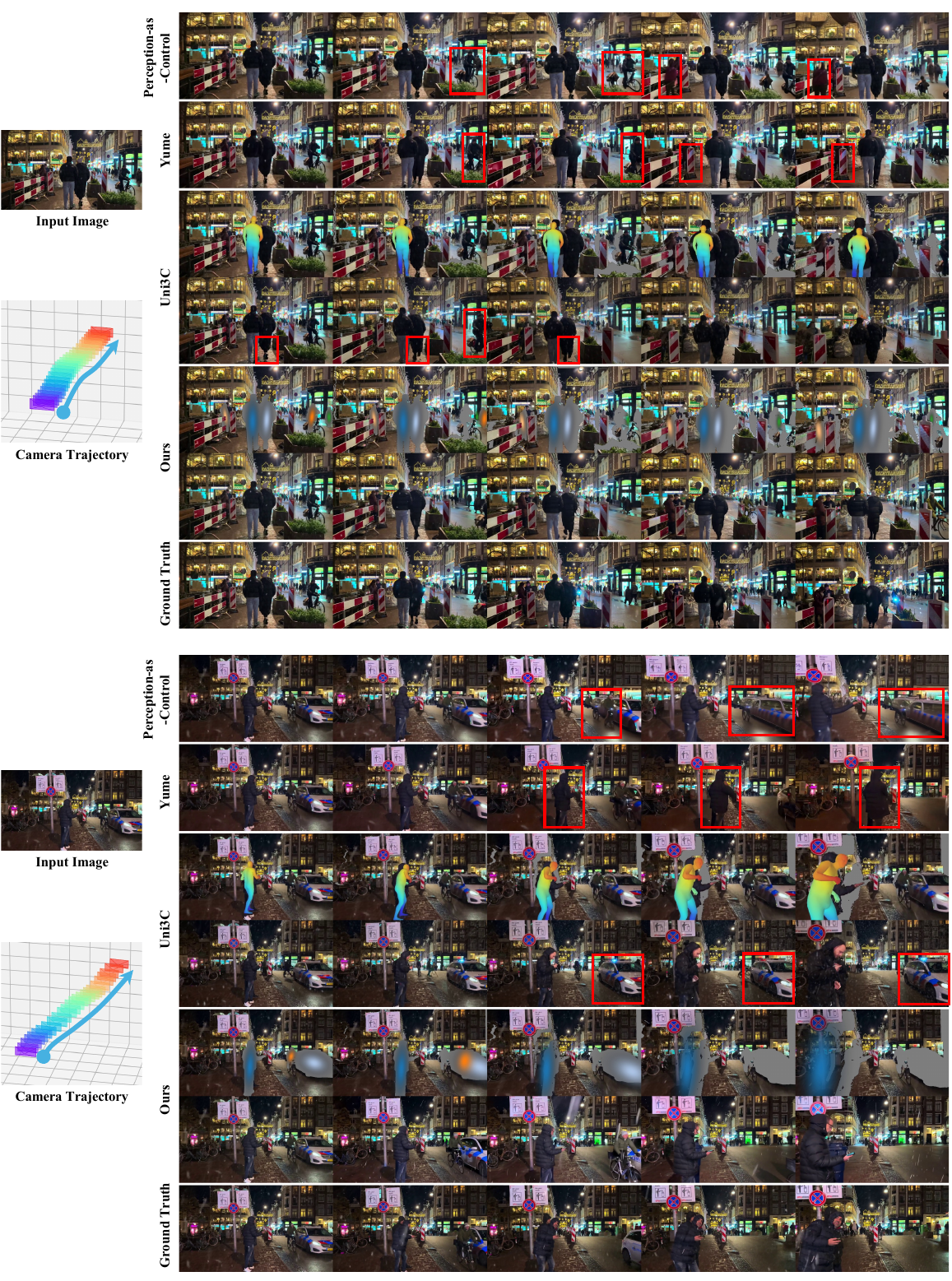}
\caption{\textbf{Additional qualitative comparison of joint camera and object motion control.}
Perception-as-Control and Uni3C exhibit noticeable object deformation, while Yume roughly follows the text-described motion but lacks precise camera control. Uni3C is also limited to a single human. In contrast, VerseCrafter more faithfully follows both the camera trajectory and multi-object motion while maintaining sharp appearance and geometrically consistent backgrounds.
\label{fig:supp_joint_control_vis2}}
\end{figure*}

We provide additional qualitative comparisons on VerseControl4D, following the same evaluation settings and baselines as in the main paper.
Fig.~\ref{fig:supp_joint_control_vis1} and Fig.~\ref{fig:supp_joint_control_vis2} present \emph{dynamic scenes} with joint camera and multi-object motion control.
Perception-as-Control and Uni3C often exhibit noticeable object deformation, while Yume roughly follows the text-described motion but lacks precise camera control.
Uni3C is also limited to a single human and does not generalize well to diverse multi-object scenarios.
In contrast, VerseCrafter more faithfully follows both the camera trajectory and multi-object motion while maintaining sharp appearance and geometrically consistent backgrounds.

\begin{figure*}
    \centering
    \includegraphics[width=1\linewidth]{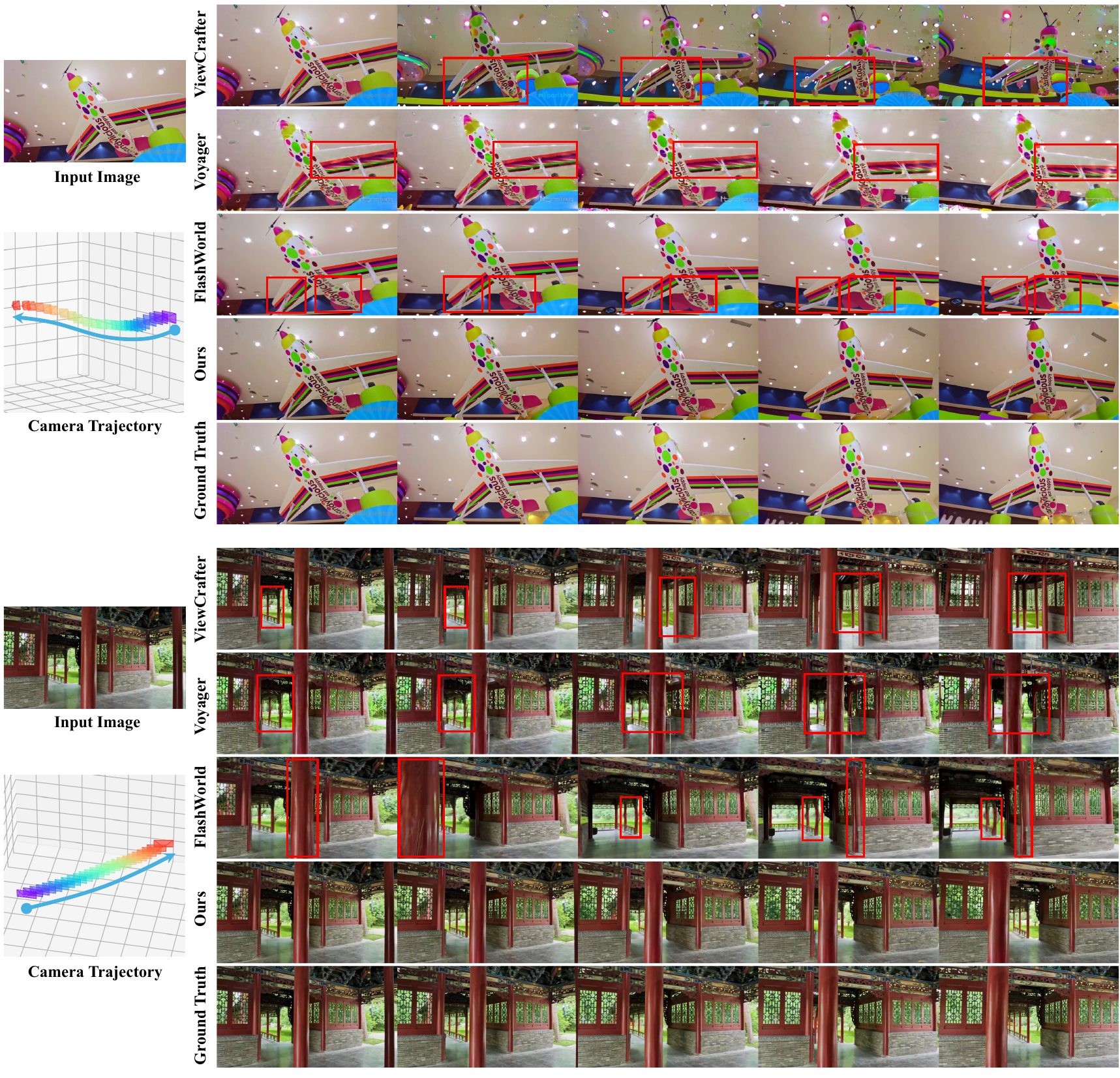}
\caption{\textbf{Additional qualitative comparison of camera-only motion control on static scenes.}
ViewCrafter, Voyager and FlashWorld exhibit distorted facades, drifting structures, or inaccurate camera motion. In contrast, VerseCrafter better follows the target camera trajectory while preserving sharp details and globally consistent 3D geometry.
\label{fig:supp_camera_only_vis1}}
\end{figure*}

\begin{figure*}
    \centering
    \includegraphics[width=1\linewidth]{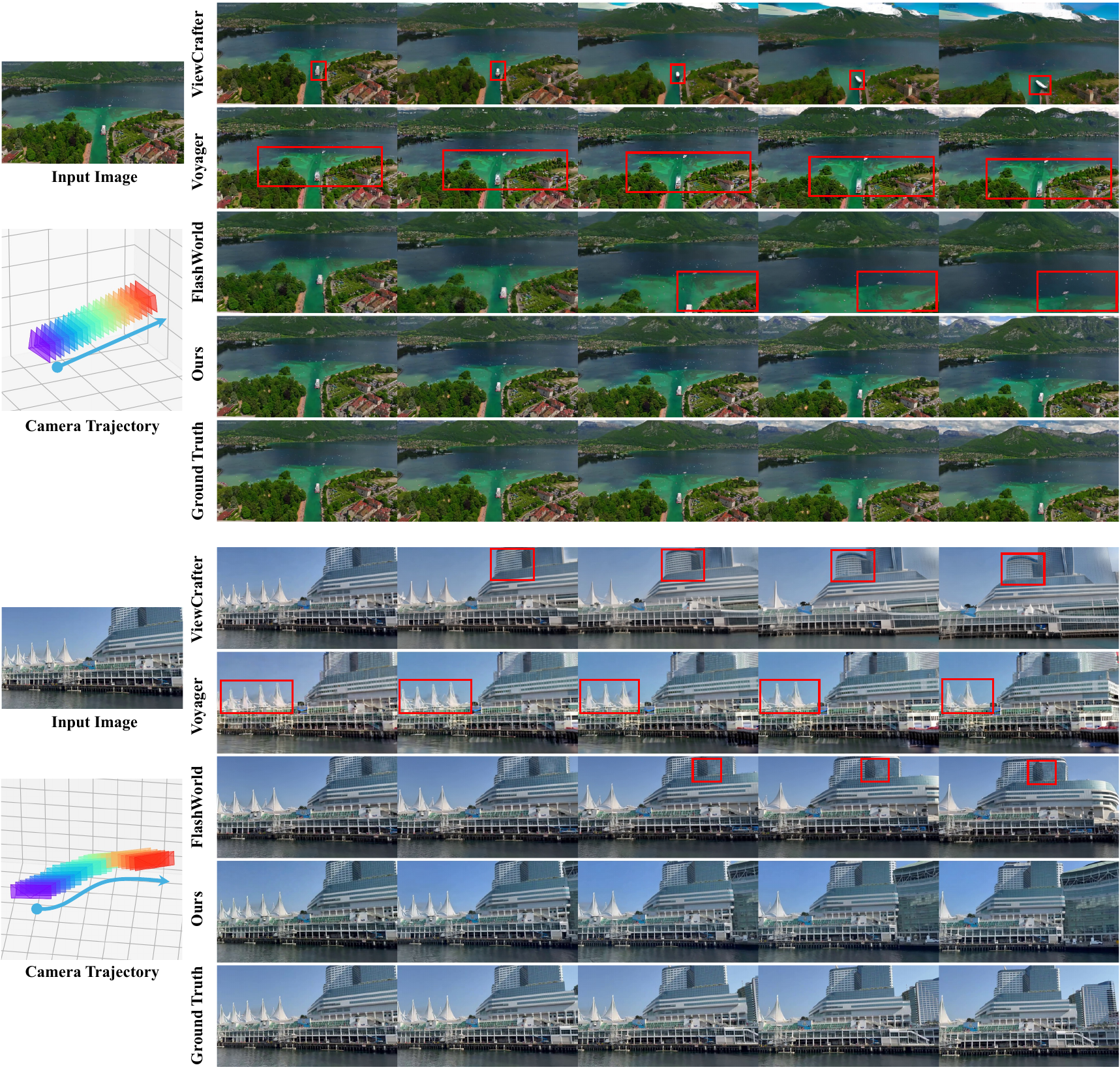}
\caption{\textbf{Additional qualitative comparison of camera-only motion control on static scenes.}
ViewCrafter, Voyager and FlashWorld exhibit distorted facades, drifting structures, or inaccurate camera motion. In contrast, VerseCrafter better follows the target camera trajectory while preserving sharp details and globally consistent 3D geometry.
\label{fig:supp_camera_only_vis2}}
\end{figure*}

Fig.~\ref{fig:supp_camera_only_vis1} and Fig.~\ref{fig:supp_camera_only_vis2} present \emph{static scenes} for camera-only motion control.
ViewCrafter, Voyager and FlashWorld often exhibit distorted facades, drifting structures, or inaccurate camera motion.
In contrast, VerseCrafter better follows the target camera trajectory while preserving sharp details and globally consistent 3D geometry.
These additional examples further demonstrate VerseCrafter's robustness under real-world 4D Geometric Control in both dynamic and static settings.

\section{Additional Analysis of Control Fidelity and Robustness}
\label{sec:additional_analysis}

We further provide targeted qualitative analyses to clarify the fidelity, scope, and robustness of our 4D Geometric Control.
Specifically, we analyze orientation controllability, dynamic background modeling, articulated and non-rigid object controllability, the effect of multi-view input, and robustness to monocular-depth errors.

\subsection{Control Fidelity and Boundary Cases}
\label{sec:control_fidelity}

\begin{figure*}[t]
    \centering
    \includegraphics[width=\linewidth]{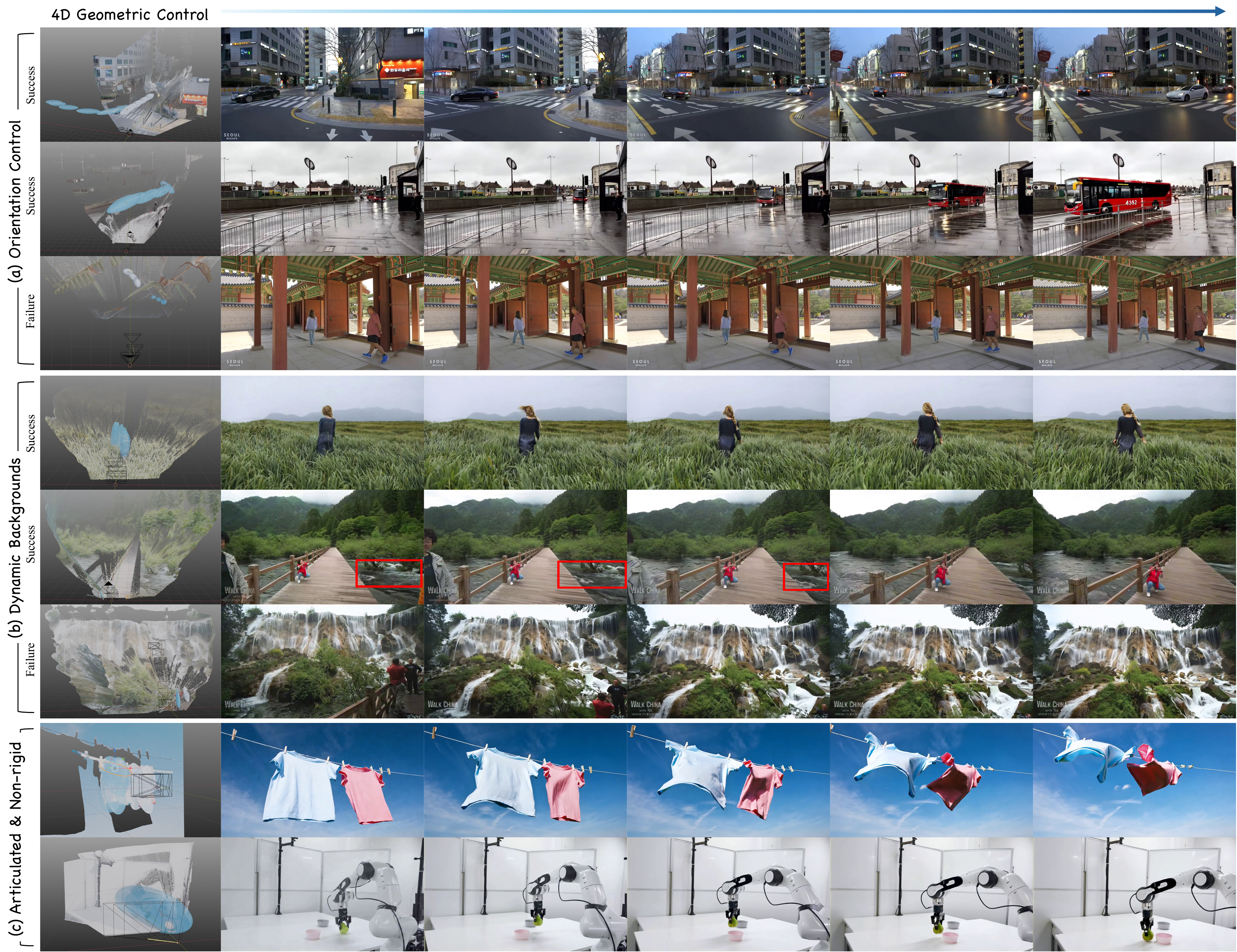}
    \caption{\textbf{Additional analysis of control fidelity and boundary cases.}
    \textbf{(a) Orientation controllability.}
    Two success cases on rigid anisotropic objects and one failure case on a human-like subject.
    \textbf{(b) Dynamic background modeling.}
    Two success cases on moderate background dynamics and one failure case on highly non-rigid background motion.
    \textbf{(c) Articulated and non-rigid object controllability.}
    Two examples showing effective coarse object-level control on articulated and non-rigid objects.}
    \label{fig:control_fidelity}
\end{figure*}

\noindent\textbf{Orientation controllability.}
Our representation provides \emph{ellipsoid-level} orientation control through the principal axes of each 3D Gaussian, rather than fine-grained 6D pose control.
As shown in Fig.~\ref{fig:control_fidelity}(a), this control is reliable for strongly anisotropic rigid objects such as cars and buses, where rotation induces clear changes in rendered footprint and depth, leading to stable orientation cues after 3D$\rightarrow$2D rendering.
However, it can fail for human-like subjects approximated by a single ellipsoid.
In such cases, heading changes mainly correspond to rotation around the ellipsoid's major principal axis, and when the other two axes are similar, the projected footprint/depth variation becomes subtle and ambiguous.
As a result, the geometric cue may be too weak to fully determine facing direction, and the diffusion prior may dominate, occasionally causing heading mismatches.

\noindent\textbf{Dynamic background modeling.}
Our background point cloud is reconstructed from the first frame and therefore serves as a mostly static geometric scaffold.
It anchors scene geometry under viewpoint changes, but does not explicitly model per-frame non-rigid background deformation.
Fig.~\ref{fig:control_fidelity}(b) shows that this design still works well for moderate dynamic-background effects such as wind-swaying grass and a flowing river, where the diffusion prior can synthesize plausible temporal variation while the 4D controls maintain camera and object consistency.
In contrast, the waterfall example exhibits weaker motion.
This failure is expected because fine, texture-dominant, highly non-rigid dynamics are only weakly constrained by a static 3D scaffold after rendering to 2D control maps.
Thus, VerseCrafter currently handles dynamic backgrounds mainly through the interaction between static geometry anchoring and the video prior, rather than through explicit background dynamics modeling.

\noindent\textbf{Articulated and non-rigid object controllability.}
VerseCrafter uses a \emph{single} 3D Gaussian per controlled object and guides its motion coarsely by changing its position, scale, and orientation over time.
This representation is not designed to explicitly encode part-level articulation.
Nevertheless, Fig.~\ref{fig:control_fidelity}(c) shows that it remains effective for object-level motion control in both articulated and non-rigid scenarios, including a robotic arm extension and wind-blown clothes.
Although the guidance is coarse, the generated videos follow the intended object-level motion while remaining visually coherent.
These examples suggest that even a simple object-level 3D Gaussian can provide a useful control signal for a broad range of dynamic objects, though finer articulation control remains an important direction for future work.

\subsection{Geometry Coverage and Robustness}
\label{sec:geometry_robustness}

\begin{figure*}[t]
    \centering
    \includegraphics[width=\linewidth]{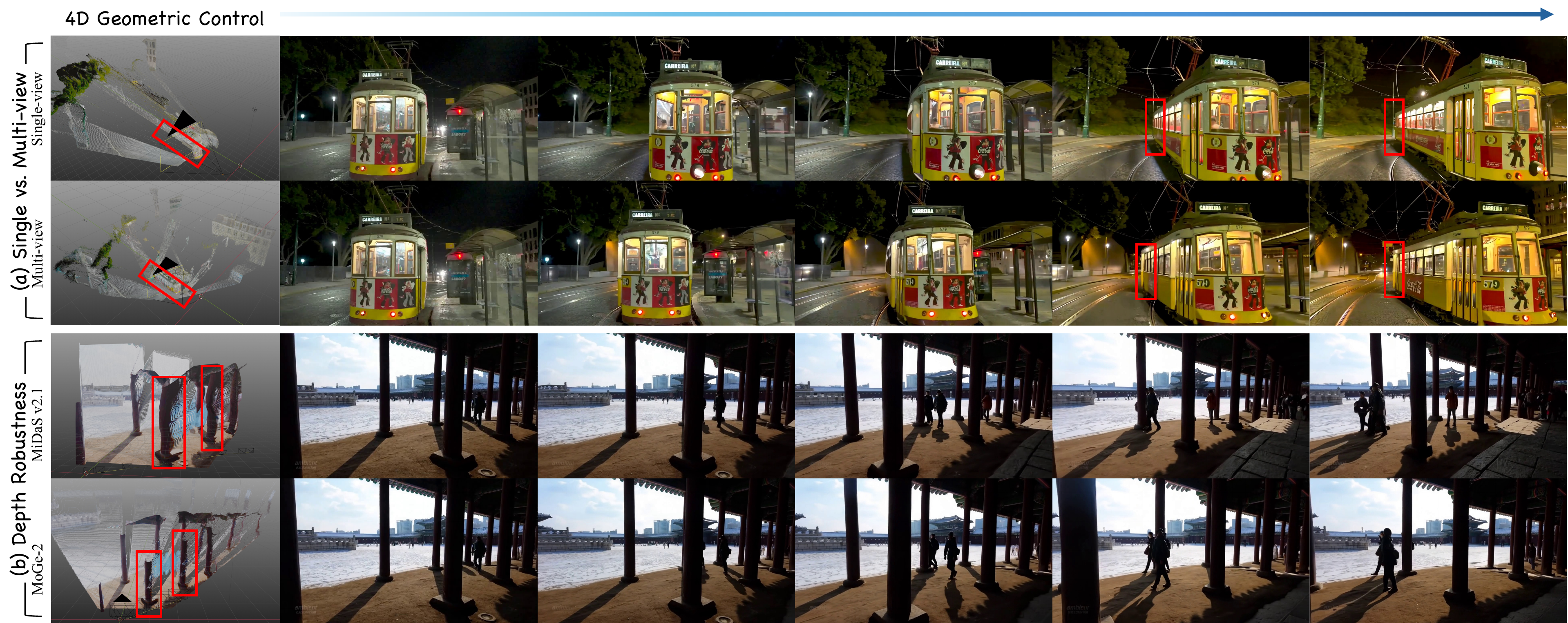}
    \caption{\textbf{Additional analysis of geometry coverage and robustness.}
    \textbf{(a) Single-view vs.\ multi-view input.}
    Multi-view reconstruction improves geometry coverage and novel-view faithfulness.
    \textbf{(b) Robustness to monocular-depth errors.}
    Even with noisier depth and distorted point clouds, the generated videos remain visually similar and preserve the main scene structure.}
    \label{fig:geometry_robustness}
\end{figure*}

\noindent\textbf{Single-view vs.\ multi-view input.}
Multi-view input improves geometry coverage and therefore improves novel-view faithfulness.
As shown in Fig.~\ref{fig:geometry_robustness}(a), using multiple views to reconstruct the scene expands the point cloud to cover regions that are weakly observed or fully invisible from a single reference image, such as the tram side and rear structure.
Consequently, the generated video is more faithful under larger viewpoint changes; for example, the rear door is recovered only in the multi-view case.
By comparison, single-view reconstruction still produces plausible videos because the diffusion prior can fill in under-constrained regions, but the missing geometry leads to less faithful novel-view synthesis.
This result supports our claim that more complete 3D reconstruction directly benefits controllable video generation when the target camera motion departs significantly from the reference view.

\setlength{\tabcolsep}{4pt}
\begin{table*}[t]
\centering
\caption{\textbf{Memory--time trade-off under different inference settings.}
We report peak per-GPU memory and diffusion inference time for the 50-step setting.
FSDP reduces peak memory from 90 GB to 70 GB with negligible runtime change, and FSDP + CPU offload further reduces it to 57 GB (36.7\% reduction) with only a small increase in diffusion inference time.}
\label{tab:fsdp_offload}
\small
\begin{tabular}{l r r r}
\toprule
\textbf{Inference setting} &
\textbf{Peak GPU memory (GB)} &
\textbf{Memory reduction (\%)} &
\textbf{Diffusion inference time (s)} \\
\midrule
Baseline                         & 90 & 0.0  & 866 \\
Baseline + FSDP                  & 70 & 22.2 & 870 \\
Baseline + FSDP + CPU offload    & 57 & 36.7 & 880 \\
\bottomrule
\end{tabular}
\end{table*}

\setlength{\tabcolsep}{4pt}
\begin{table}[t]
\centering
\caption{\textbf{Stage-wise end-to-end inference latency and cacheability.}
We report the runtime breakdown for generating an 81-frame 720P video on 8$\times$96GB GPUs.
4D geometric scene state is reusable across repeated edits of the same scene, and model loading is a one-time startup cost, whereas 4D control rendering and diffusion sampling must be rerun when the edited controls change.}
\label{tab:inference_time}
\small
\begin{tabular}{lcc}
\toprule
\multicolumn{1}{c}{\textbf{Stage}}       & \textbf{Time (s) $\downarrow$} & \textbf{Cacheable?}   \\
\midrule
4D Geometric State Construction                    & $\sim$23              & \cmark \\
4D Control Maps Rendering                     & $\sim$60              & \xmark \\
\textbf{Diffusion Sampling}              &                        &                       \\
\quad\quad Diffusion Model Loading                      & $\sim$203             & \cmark \\
\quad\quad Diffusion Inference                & $\sim$866 (50 steps)  & \xmark \\
\quad\quad Diffusion Inference                & $\sim$715 (30 steps)  & \xmark \\
\bottomrule
\end{tabular}
\end{table}

\noindent\textbf{Robustness to monocular-depth errors.}
We also test robustness to imperfect monocular depth estimation in challenging conditions with heavy occlusion and strong illumination variation.
In Fig.~\ref{fig:geometry_robustness}(b), replacing MoGe-2 with MiDaS v2.1 produces visibly noisier depth and distorted point clouds in difficult regions such as the pillars highlighted by the red boxes.
Despite these geometry errors, the generated videos remain visually similar and preserve the main building structure.
This robustness is expected because the reconstructed point cloud acts as a \emph{coarse} geometric scaffold rather than a per-pixel hard constraint.
After 3D$\rightarrow$2D rendering, the diffusion prior can compensate for moderate depth noise and still generate structurally plausible results.
Therefore, while better monocular geometry generally improves controllability, VerseCrafter does not critically depend on perfectly accurate depth estimates.

\noindent
Overall, these analyses suggest that VerseCrafter is most effective when the underlying 4D geometric cues are sufficiently informative, while failures mainly arise in under-constrained cases such as subtle human orientation changes or highly non-rigid background dynamics.

\section{Inference Efficiency and Memory Usage}

We further analyze the inference cost of VerseCrafter.
For generating an 81-frame 720P video on 8$\times$96GB GPUs, Table~\ref{tab:inference_time} shows that diffusion inference is the dominant bottleneck, while 4D geometric state construction is cacheable across repeated edits of the same scene and diffusion model loading is a one-time startup cost.
Accordingly, the per-edit latency is substantially reduced for subsequent edits, and can be further lowered by using fewer denoising steps.

Table~\ref{tab:fsdp_offload} summarizes the memory and runtime trade-off under different inference settings.
FSDP substantially reduces peak per-GPU memory with negligible runtime overhead, and FSDP + CPU offload further lowers memory at only a small additional cost.
These results suggest that the current practical bottleneck is diffusion inference rather than 4D geometric state construction.

\section{Limitations and Future Work}
\label{sec:supp_limitations}

Despite the encouraging results, VerseCrafter still has several limitations that suggest promising directions for future work.

First, our current object representation provides only \emph{ellipsoid-level} control through a single 3D Gaussian per object, which limits fine-grained pose and part-level articulation, especially for human-like or near-symmetric objects. More expressive object representations, such as multiple Gaussians per object or articulated 3D structures, may improve fine-grained orientation and pose control.

Second, our background point cloud is reconstructed from the first frame and serves as a mostly static geometric scaffold, which limits controllability for highly non-rigid and texture-dominant scene dynamics such as waterfalls. Incorporating explicit dynamic background representations or temporally evolving scene geometry may improve controllability in such cases.

Third, although VerseCrafter enforces 4D geometric consistency through explicit camera control and 3D Gaussian trajectory control, it does not impose explicit physical constraints during generation. Integrating stronger physics priors, such as collision-aware losses, contact constraints, ground constraints, or differentiable physics guidance, could improve physical realism and controllability in complex interactions.

Finally, VerseCrafter remains computationally expensive at high resolution and long temporal horizons because it conditions a large frozen video diffusion backbone and renders multi-channel 4D controls for all frames. Future work may explore more efficient backbones, distilled sampling, cached control encoding, and streaming or long-video synthesis to enable faster and longer world rollouts.

{
    \small
    \bibliographystyle{ieeenat_fullname}
    \bibliography{main}
}

\end{document}